\documentclass[lettersize,journal]{IEEEtran}
\usepackage{amsmath,amsfonts}
\usepackage{algorithmic}
\usepackage{algorithm}
\usepackage{array}
\usepackage[caption=false,font=footnotesize,labelfont=sf,textfont=sf]{subfig}
\usepackage{textcomp}
\usepackage{stfloats}
\usepackage{url}
\usepackage{verbatim}
\usepackage{graphicx}
\usepackage{cite}
\usepackage{booktabs}
\usepackage{multirow}
\usepackage[abs]{overpic}
\usepackage{hyperref}
\usepackage{color}
\usepackage{colortbl}
\usepackage{xcolor}
\definecolor{lightgray}{gray}{0.7}
\makeatletter

\begin{document}

\title{OBMO: One Bounding Box Multiple Objects \\for Monocular 3D Object Detection}

\author{Chenxi Huang, Tong He, Haidong Ren, Wenxiao Wang, Binbin Lin$^{\ast}$\thanks{* Corresponding author}, Deng Cai\textit{, Member, IEEE}
\thanks{C. Huang and D. Cai are with the State Key Laboratory of CAD\&CG, College of Computer Science, Zhejiang University, Hangzhou, Zhejiang 310058, China (emails: hcx\_98@zju.edu.cn, dengcai@cad.zju.edu.cn).}
\thanks{T. He is with Shanghai AI Laboratory, Shanghai, China (emails: tonghe90@gmail.com).}
\thanks{H. Ren is with Ningbo Zhoushan Port Group Co.,Ltd., Ningbo, China. (email: rhdong@nbport.com.cn).}
\thanks{W. Wang and B. Lin are with College of Software, Zhejiang University, Ningbo, Zhejiang, China (emails: wenxiaowang@zju.edu.cn, binbinlin@zju.edu.cn).}
}
\markboth{Journal of \LaTeX\ Class Files}%
{Shell \MakeLowercase{\textit{et al.}}: A Sample Article Using IEEEtran.cls for IEEE Journals}

% \IEEEpubid{0000--0000/00\$00.00~\copyright~2022 IEEE}
% Remember, if you use this you must call \IEEEpubidadjcol in the second column for its text to clear the IEEEpubid mark.

\maketitle
\begin{abstract} % 150-250 words
Compared to typical multi-sensor systems, monocular 3D object detection has attracted much attention due to its simple configuration. 
However, there is still a significant gap between LiDAR-based and monocular-based methods. 
In this paper, we find that the ill-posed nature of monocular imagery can lead to depth ambiguity. Specifically, objects with different depths can appear with the same bounding boxes and similar visual features in the 2D image. Unfortunately, the network cannot accurately distinguish different depths from such non-discriminative visual features, resulting in unstable depth training.
To facilitate depth learning, we propose a simple yet effective plug-and-play module, \underline{O}ne \underline{B}ounding Box \underline{M}ultiple \underline{O}bjects (OBMO).
Concretely, we add a set of suitable pseudo labels by shifting the 3D bounding box along the viewing frustum. To constrain the pseudo-3D labels to be reasonable, we carefully design two label scoring strategies to represent their quality.
In contrast to the original hard depth labels, such soft pseudo labels with quality scores allow the network to learn a reasonable depth range, boosting training stability and thus improving final performance. 
Extensive experiments on KITTI and Waymo benchmarks show that our method significantly improves state-of-the-art monocular 3D detectors by a significant margin (The improvements under the moderate setting on KITTI validation set are $\mathbf{1.82\sim 10.91\%}$ \textbf{mAP in BEV} and $\mathbf{1.18\sim 9.36\%}$ \textbf{mAP in 3D}). Codes have been released at \url{https://github.com/mrsempress/OBMO}.
\end{abstract}
\begin{IEEEkeywords}
3D object detection, Monocular images, Depth ambiguity, Camera project principles.
\end{IEEEkeywords}

%%%%%%%%% BODY TEXT
\section{Introduction}
\label{sec:intro}
\IEEEPARstart{D}{ue} to widely deployed applications in robot navigation and autonomous driving~\cite{vision,computer,mv3D,rey2002automatic,deepdriving}, 3D object detection has become an active research area in computer vision.
	Although LiDAR-based 3D object detectors~\cite{xu2021spg,xu2021behind,sessd} have achieved excellent performance because of accurate depth measurements, the application of these methods is still constrained by the high cost of 3D sensors, limited working range, and sparse data representation.
    Monocular-based 3D detectors~\cite{bao2019monofenet,park2021pseudo,liu2021learning,MonoJSG,MonoDTR, bui2021gac3d}, on the other hand, have received increasing attention in autonomous driving due to their easy accessibility and rich semantic clues.
	
\begin{figure}[t]
  \centering
   \includegraphics[width=0.95\linewidth]{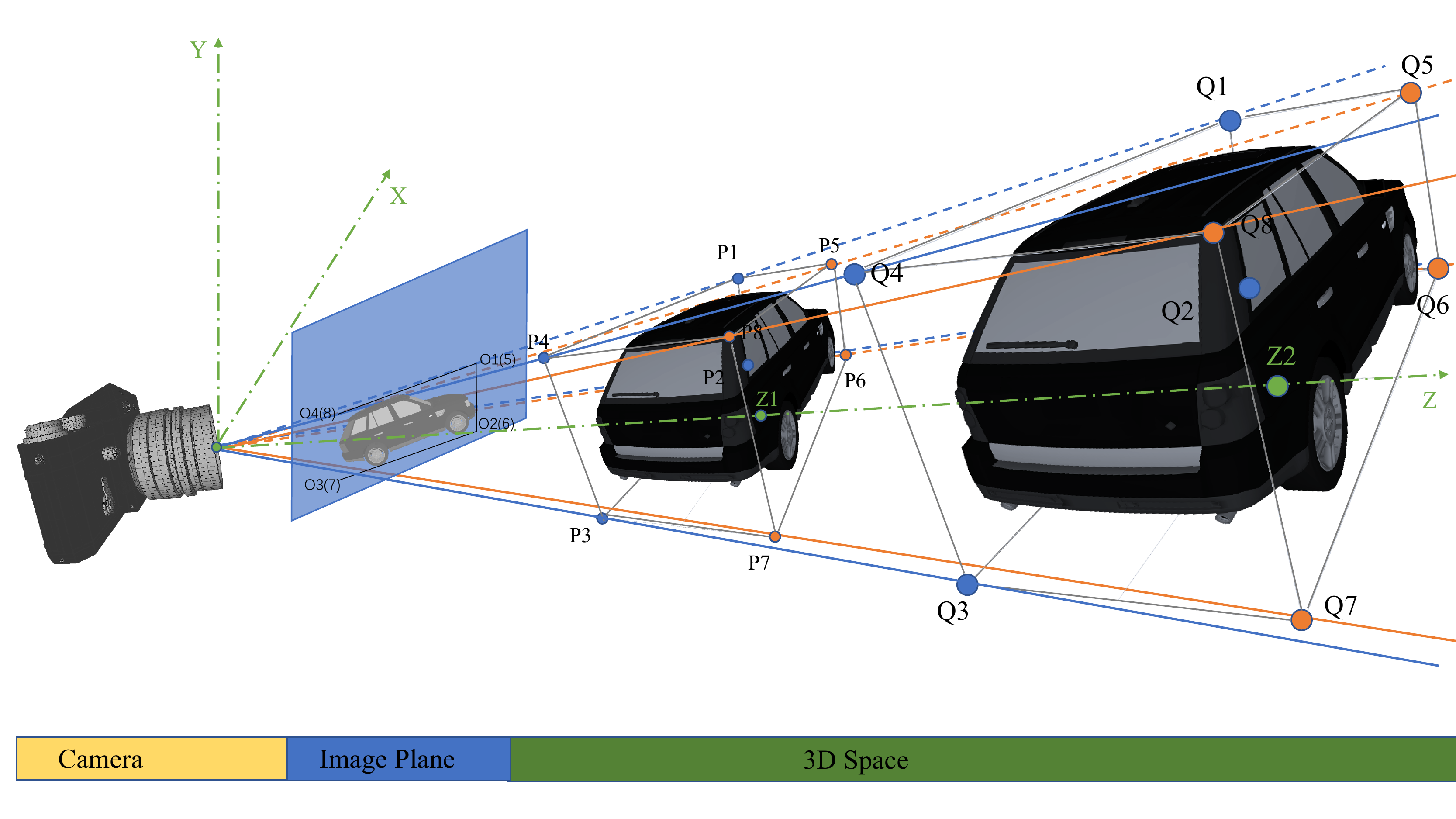}
   \caption{
   Objects with different depths and dimensions in 3D space.
   Objects $P$ and $Q$ have the same bounding box and similar visual features in the 2D image, leading to depth ambiguity.
   }
   \label{Fig: camera}
\end{figure}

	Though tremendous efforts have recently been devoted to improving the accuracy, monocular-based 3D object detection is still highly challenging, as substantiated by ~\cite{bui2021gac3d,beker2020monocular,wang2021progressive,nobis2020exploring}.
    Earlier works are based on mature 2D object detection, using 2D Region of interests (ROIs) to regress 3D information.
    \cite{M3D-RPN,Deep3DBox,Mono3D,Mono3D++,MonoGRNet} follow the pipeline by either introducing geometrical priors or laying 2D-3D constraints.
    Promising results have been achieved; however, the gap in the accuracy between LiDAR-based and monocular-based approaches is still significant.
    One of the critical reasons for the less competitiveness in monocular-based methods is lacking precise knowledge of depth.\looseness=-1
    
    To this end, many prior works~\cite{CaDDN,MonoFlex,Monodle,Homographyloss,PatchNet,Pseudo-LiDAR,Pseudo-LiDAR++,shiftrcnn,decoupled3d,monopair,GUPNet,MonoRCNN,Ground-aware} focus on improving the accuracy of instance depth estimation.
    \IEEEpubidadjcol
    These methods mainly involve two strategies to encode depth prior by either building dependencies on the intermediate task of depth prediction or adding geometric constraints on the final results. 
    For the former, the expression content of the data is enriched by initial depth prediction values from monocular depth estimation or extra-designed modules. Some methods transform the front view into other views using the predicted depth values, such as the bird's eye view (BEV)~\cite{CaDDN,Homographyloss}. Other methods combine the predicted depth values with the corresponding RGB values into new data representations, for example, concatenating them on channels~\cite{PatchNet} or converting them into LiDAR format~\cite{Pseudo-LiDAR,Pseudo-LiDAR++}.
    For the latter, they add extra modules or change the objective function to assist in estimating depth. 
    \cite{shiftrcnn,Ground-aware} use the projection relationship to constrain the predictions by well-designed Volume Displacement loss and ground-aware convolution module, respectively. 
    \cite{monopair,GUPNet,MonoRCNN} split the depth value into a coarse value and a bias-corrected value.
    \cite{monopair} considers the relationship between objects and regards the distance of the 3D pair as bias. \cite{GUPNet,MonoRCNN} predict the variance of depth and height, respectively, and use them to fine-tune the rough values.
    
    Previous works adopt the one-to-one learning strategy, which uses a 3D object label to supervise learning from the visual features of one 2D object. However, due to the asymmetrical projection between 2D and 3D,
    this one-to-one learning strategy often causes depth ambiguity.
    For example, if moving a car along the viewing frustum from a depth of $50$~m to $55$~m and enlarging the size by $1.1$, the visual representation remains unchanged when projected to the 2D plane.
    As shown in Figure~\ref{Fig: camera}, different objects in 3D space may have very similar bounding boxes and visual features when projected to the 2D image.
    Considering that the average length, width, and height of the Car are $3.88$~m, $1.63$~m, and $1.53$~m, respectively, in KITTI, it is still maintained in a reasonable scope when the size expanded by $1.1$ times. 
    Consequently, such ambiguity often causes inferior performance the network has to distinguish different depths based on the non-discriminative visual clues. 
    
	We propose a simple yet effective plug-and-play module named
	\underline{\textbf{O}}ne \underline{\textbf{B}}ounding Box \underline{\textbf{M}}ultiple \underline{\textbf{O}}bjects (OBMO) to address the above problems.
	The core idea of OBMO is adding reasonable pseudo labels by shifting the depth of an object along the viewing frustum. 
    Considering the lack of depth in 2D images, the soft pseudo labels of 3D objects play a significant role in encoding depth prior. Compared with the hard labels, such soft labels encourage the network to learn the depth distribution and stabilize the learning process, due to the less variance in the gradient between training cases~\cite{Distilling}.

    Designing such soft labels is non-trivial, as the significant variation of depth often generates invalid sizes of 3D objects, making the network overwhelmed by negative samples. To this end, we design two label scoring strategies that use dimensional priors and geometric constraints to represent the quality of pseudo labels.
	
	By introducing the OBMO module and the label scoring strategy, the one-to-many problem is addressed to some extent: the network is encouraged to learn a soft distribution of object locations rather than deterministic ones.
	To show the superiority of OBMO, we perform extensive experiments on KITTI and Waymo datasets. 
	Multiple monocular 3D detectors are used, 
	including direct regression-based detectors like RTM3D~\cite{RTM3D}, Ground-aware~\cite{Ground-aware}, GUPNet~\cite{GUPNet} and depth-aware detectors like PatchNet~\cite{PatchNet}, Pseudo-LiDAR~\cite{Pseudo-LiDAR}. 
	Experimental results show that our method stabilizes the training process and improves the overall BEV and 3D detection performance, as shown in Figure ~\ref{Fig: depth}.
	Concretely, on the widely used KITTI dataset, our approach significantly improves the state-of-the-art (SOTA) monocular 3D detectors by $\mathbf{1.82\sim 10.91\%}$ \textbf{mAP in BEV} and $\mathbf{1.18\sim 9.36\%}$ \textbf{mAP in 3D}. On the larger Waymo open dataset, we boost GUPNet with $\mathbf{3.34\%}$ \textbf{mAP} gains under the \text{LEVEL 1} ($\text{IoU}=0.5$) setting. \looseness=-1

The contributions can be summarized as follows:
\begin{itemize}
\item We point out that the depth ambiguity problem in monocular 3D detection 
has been ignored
in previous methods and argue that this problem can result in unstable depth training, which undermines performance.
\item To alleviate the problem of unstable depth training in monocular 3D object detection, we propose a plug-and-play module OBMO.
It explicitly adds a set of suitable pseudo labels by 
shifting bounding boxes along the viewing frustum
for each original object.
\item We design two label scoring strategies to represent the qualities of pseudo labels: IoU Label Scores and Linear Label Scores, which are inspired by the fixed dimension range of objects in the same category.
\item We conduct extensive experiments on various datasets: KITTI and Waymo. The consistent improvement of the accuracy demonstrates the effectiveness of our proposed OBMO. For example, we achieved 21.41\% in $AP_{BEV}$ and 15.70\% in $AP_{3D}$ under the moderate KITTI validation set based on GUPNet, improving the state-of-the-art results substantially.
\end{itemize}

%-------------------------------------------------------------------------
\section{Related Work}
\label{sec:rel}
%-------------------------------------------------------------------------
\subsection{LiDAR 3D Object Detection}
Due to the accurate depth measurement, most state-of-the-art 3D object detection methods are based on LiDAR~\cite{3dssd,pointpillars,pvrcnn,guo2021pct,sessd}.
These methods can be roughly divided into two parts: voxel-based methods and point-based methods. 

\subsubsection{voxel-based methods}
In order to tackle the irregular data format of point clouds, voxel-based methods~\cite{mv3D,voxelnet,voxelrcnn} convert the irregular point clouds into regular voxel grids. Then, use mature convolution neural architectures to extract high-level features.
However, the receptive fields are constrained by the kernel size of 2D/3D convolutions~\cite{peng2017large,guo2022visual}. Moreover, the computation and memory grow cubically with the input resolution. 
To this end, SECOND~\cite{second} leverages the 3D submanifold sparse convolution. In spatially sparse convolution, output points are not computed if there is no related input point, which signiﬁcantly increases the speed of both training and inference.
Further, PointPillars~\cite{pointpillars} is proposed to simplify the voxels to pillars.
Overall, voxel-based methods can achieve good detection performance with promising efficiency. 
However, it is difficult to determine the optimal voxel resolution in practice since the complex geometry and various dimension objects.

\subsubsection{point-based methods}
Point-based methods~\cite{pointrcnn,pointgnn} directly extract raw unstructured point cloud features via different set abstraction operations. Further, it generates specific proposals for objects of interest. 
These point-based methods, such as the PointNet~\cite{pointnet} series, enable ﬂexible receptive ﬁelds for point cloud feature learning.
For example, PointRCNN~\cite{pointrcnn}, a two-stage 3D region proposal framework for 3D object detection, generates object proposals from segmented foreground points and exploits semantic features to regress high-quality 3D bounding boxes.
PointGNN~\cite{pointgnn} generalizes graph neural networks to do 3D object detection.
In conclusion, point-based methods don't need extra preprocessing steps such as voxelization. However, the main bottleneck of point-based methods is insufficient representation and inefficiency.

\subsection{Monocular 3D Object Detection}
Although LiDAR 3D object detectors present promising results, 
they have disadvantages of the limited working range and sparse data representation. 
Monocular 3D object detectors, on the other hand, enjoy the low cost and high frame rate. 
Current monocular 3D object detection methods can be roughly divided into two categories: direct regression-based methods and depth-aware methods. 

\subsubsection{Direct Regression-based Methods}
Direct regression-based methods~\cite{bao2019monofenet,park2021pseudo,liu2021learning} obtain the 3D detection results directly from RGB images without extra knowledge like depth maps, stereo images, etc.

Mono3D~\cite{Mono3D} first proposes an energy minimization approach and assumes that all vehicles are placed on the ground plane. Moreover, it scores each candidate box projected to the image plane via several intuitive potentials encoding semantic segmentation, contextual information, size and location priors, and typical object shape.
Deep3DBox~\cite{Deep3DBox} simplifies the whole pipeline by removing extra 3D shape models and complex pre-processing operators.
It is based on 2D object detection and uses geometric constraints that the 3D bounding box should fit tightly into the 2D detection bounding box. 
Considering geometric reasoning, MonoGRNet~\cite{MonoGRNet} simultaneously estimates 2D bounding boxes, instance depth, 3D location of objects, and local corners. 
M3D-RPN~\cite{M3D-RPN} proposed depth-aware convolutional layers for learning spatially-aware features to produce 3D proposals directly. 

SMOKE~\cite{SMOKE} removes the 2D detection part and directly estimates 3D position by predicting projected 3D centers. 
RTM3D~\cite{RTM3D} adds eight corner points as keypoints so that 
more geometric constraints can be applied to remove false alarms. 
It also designs a keypoint feature pyramid, which uses soft weight by a softmax operation to denote the importance of each scale. 
Center3D~\cite{Center3D} uses Linear Increasing Discretization and a combination of classification and regression branches to predict depth. 
MonoFlex~\cite{MonoFlex} explicitly decouples the truncated objects and adaptively combines multiple approaches for object depth estimation. Specifically, it divides objects according to whether their projected centers are “inside” or “outside” the image. Furthermore, it formulates the object depth estimation as an uncertainty-guided ensemble of directly regressed object depth and solved depths from different groups of keypoints.
GUPNet~\cite{GUPNet} proposes a GUP module to obtain the geometry-guided uncertainty of the inferred depth and designs a Hierarchical Task Learning strategy to reduce the instability caused by error amplification. 
MonoDTR~\cite{MonoDTR} combines the transformer architecture and proposes a Depth-Aware Transformer module, which is used to integrate context- and depth-aware features globally.

Direct regression-based methods predict depth through a branch and employ one depth value to supervise a Region of Interest (ROI). However, we argue that this one-to-one learning strategy often suffers from depth ambiguity problems
in monocular 3D object detection.

%-------------------------------------------------------------------------
\subsubsection{Depth-aware Methods}
Depth-aware methods usually need extra depth map, which is used for 3D detection.

Pseudo-LiDAR~\cite{Pseudo-LiDAR} combines monocular 3D object detection task with monocular depth estimation task. It transforms RGB images to point clouds via an off-the-shelf depth estimator. Finally, effective point cloud-based 3D object detectors are employed for achieving the detection results. 
PatchNet~\cite{PatchNet} discovers that the data representation is not the most important one, but the coordinate transformation is. Thus it directly integrates the 3D coordinates as additional channels of RGB image patches. 
D4LCN~\cite{D4LCN} points out that methods like Pseudo-LiDAR highly rely on the quality of depth map, and traditional 2D convolution cannot distinguish foreground pixels and background pixels. So it generates dynamic convolution kernels to extract features in different 3D locations. 
CaDDN~\cite{CaDDN} discretizes the range of depth and utilizes estimated categorical pixel-wise depth distribution. It changes the representation into BEV and then uses the BEV backbone to predict the 3D detection results. 
MonoJSG~\cite{MonoJSG} reformulates the Monocular Object Depth Estimation as a progressive refinement problem and proposes a joint semantic and geometric cost volume to model the depth error.

Depth-aware methods only obtain a single depth value of the center point pixel through the Monocular Depth Estimation task. Similarly, they ignore the possibility of multiple reasonable depth values.

%-------------------------------------------------------------------------
\section{Approach}
\label{sec:app}
In this section, we first provide a detailed analysis of the widespread existence of ``one bounding box with multiple objects." 
Such ambiguity severely affects the training stability and accuracy of the model. 
Previous works ignore this problem, while we propose a simple but efficient module OBMO to lessen the impact.
Since the dimension of each category has its reasonable range, we design two label scoring strategies to represent the quality of pseudo labels, making unreasonable pseudo labels ineffective.

\subsection{Depth Ambiguity Problem}
\label{Sec: analyze}
%-------------------------------------------------------------------------
Obviously, 3D space is much larger than the projected 2D space. Using a 2D image to recover 3D space is an ill-posed task. 
Considering two objects with different 3D locations in 3D space, they may have similar bounding boxes and visual features in the 2D image, as shown in Figure~\ref{Fig: camera}. It indicates that predicting precise 3D locations from the 2D image may be impossible. We theoretically prove it in this subsection.

Without loss of generality, we assume that the camera system has been calibrated, which follows a typical pinhole imaging principle, as shown in \text{Equation~\ref{Eq: principles}}.

\begin{equation}
\begin{aligned}
s\left[\begin{matrix} u \\ v \\ 1 \end{matrix} \right]_{3\times 1}=&\left[\begin{matrix} f_{x} & 0 &c_{x}& 0 \\ 0 & f_{y} & c_{y} &  0\\0 & 0 & 1 & 0 \end{matrix} \right]_{3\times 4} \left[\begin{matrix}x \\ y \\ d \\ 1\end{matrix} \right]_{4\times 1}
\end{aligned}
\label{Eq: principles}
\end{equation}

In this equation, $s$ is the scale factor, $u$, $v$ represent the position of an object in image coordinates, $x$, $y$ refer to its position in camera coordinates, $d$ is the depth of the object. $f_{x}$, $f_{y}$, $c_{x}$, $c_{y}$ come from intrinsic parameters of the calibrated camera. We set the scale factor $s=1$ for notation convenience.
Then, we can rewrite Equation~\ref{Eq: principles} as follows:

\begin{equation}
\begin{aligned}
\frac{u - c_{x}}{f_{x}} = \frac{x}{d}, \qquad \frac{v - c_{y}}{f_{y}} =  \frac{y}{d}
\end{aligned}
\label{Eq: simple}
\end{equation}

Regarding a point $A (u,v)$ on the image, $\frac{u - c_{x}}{f_{x}}$ and $\frac{v - c_{y}}{f_{y}}$ are fixed, as $f_{x}$, $f_{y}$, $c_{x}$, $c_{y}$ are intrinsic parameters of the camera. 
According to Equation~\ref{Eq: simple}, $\frac{u - c_{x}}{f_{x}}$ denotes the ratio between $x$ and $d$, while $\frac{v - c_{y}}{f_{y}}$ denotes the ratio between $y$ and $d$. Therefore, we call them \text{X-Z} ratio and \text{Y-Z} ratio, respectively.
According to the projection relationship, we know that infinite 3D points can lead to $A$, as long as they have the same \text{X-Z} ratio and \text{Y-Z} ratio (along the same ray from the camera optical center to the 3D point $(x,y,d)$).
As a result, one point on the image can correspond to multiple 3D locations, and one 2D bounding box on the image can correspond to various objects in 3D space.

\begin{figure}[t]
  \centering
  \subfloat[\scriptsize{3D view with $\text{yaw}=0$.}]{\includegraphics[width=0.47\linewidth]{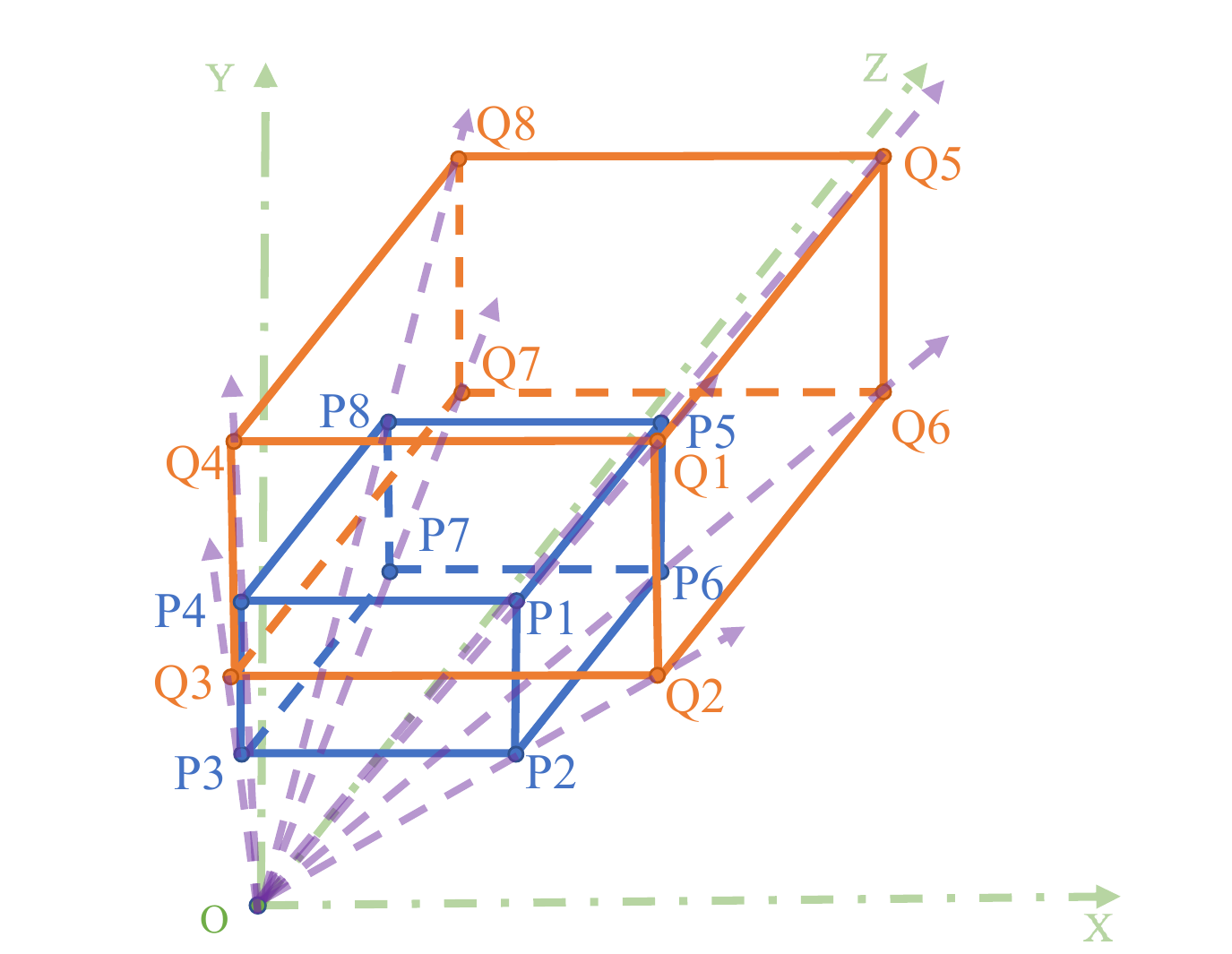}
  \label{fig:font}}
  \hfill
  \subfloat[\scriptsize{BEV with $\text{yaw}=0$.}]{\includegraphics[width=0.47\linewidth]{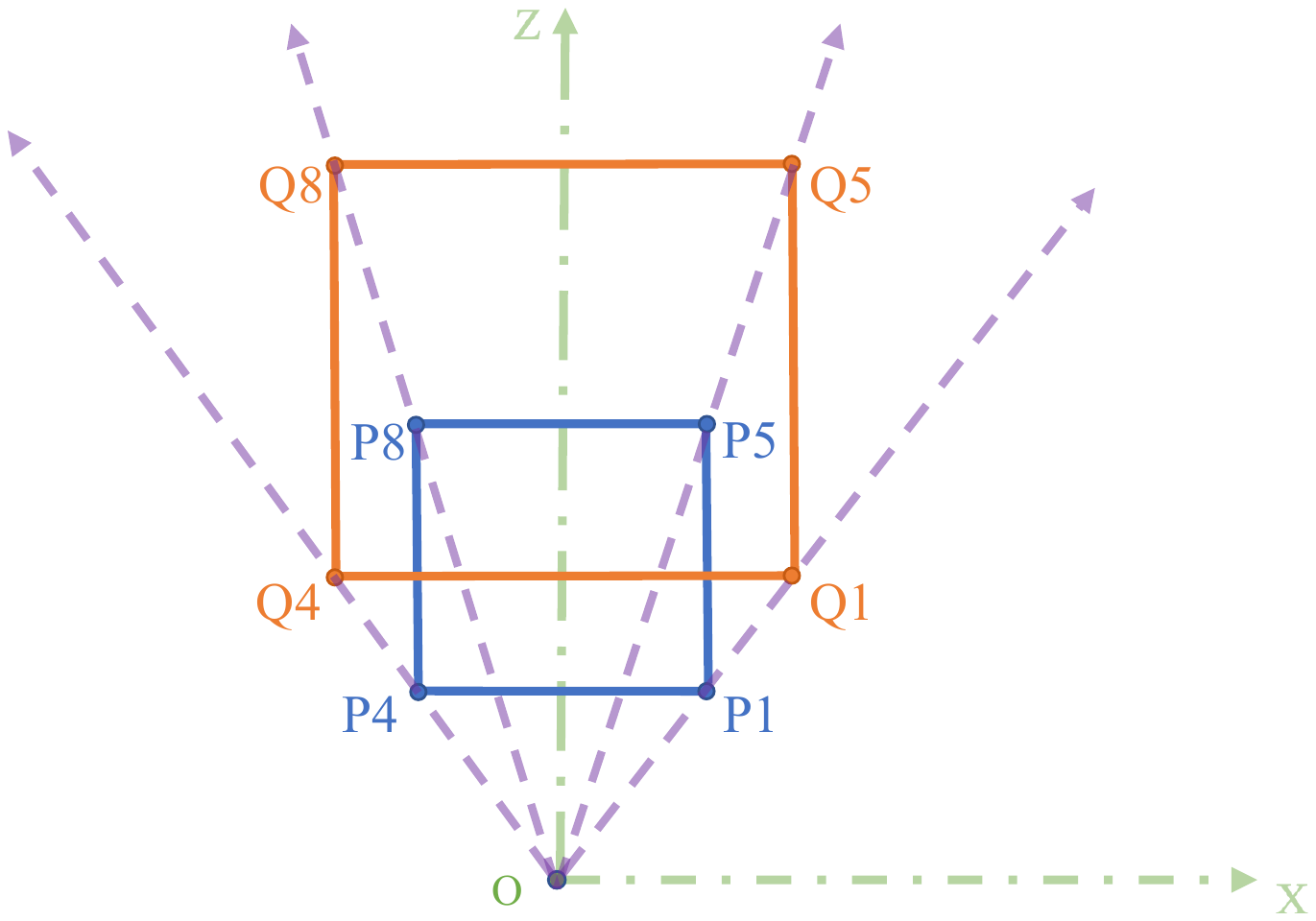}
  \label{fig:BEV}}
  \\
  \subfloat[\scriptsize{Lateral view with $\text{yaw}=0$.}]{\includegraphics[width=0.47\linewidth]{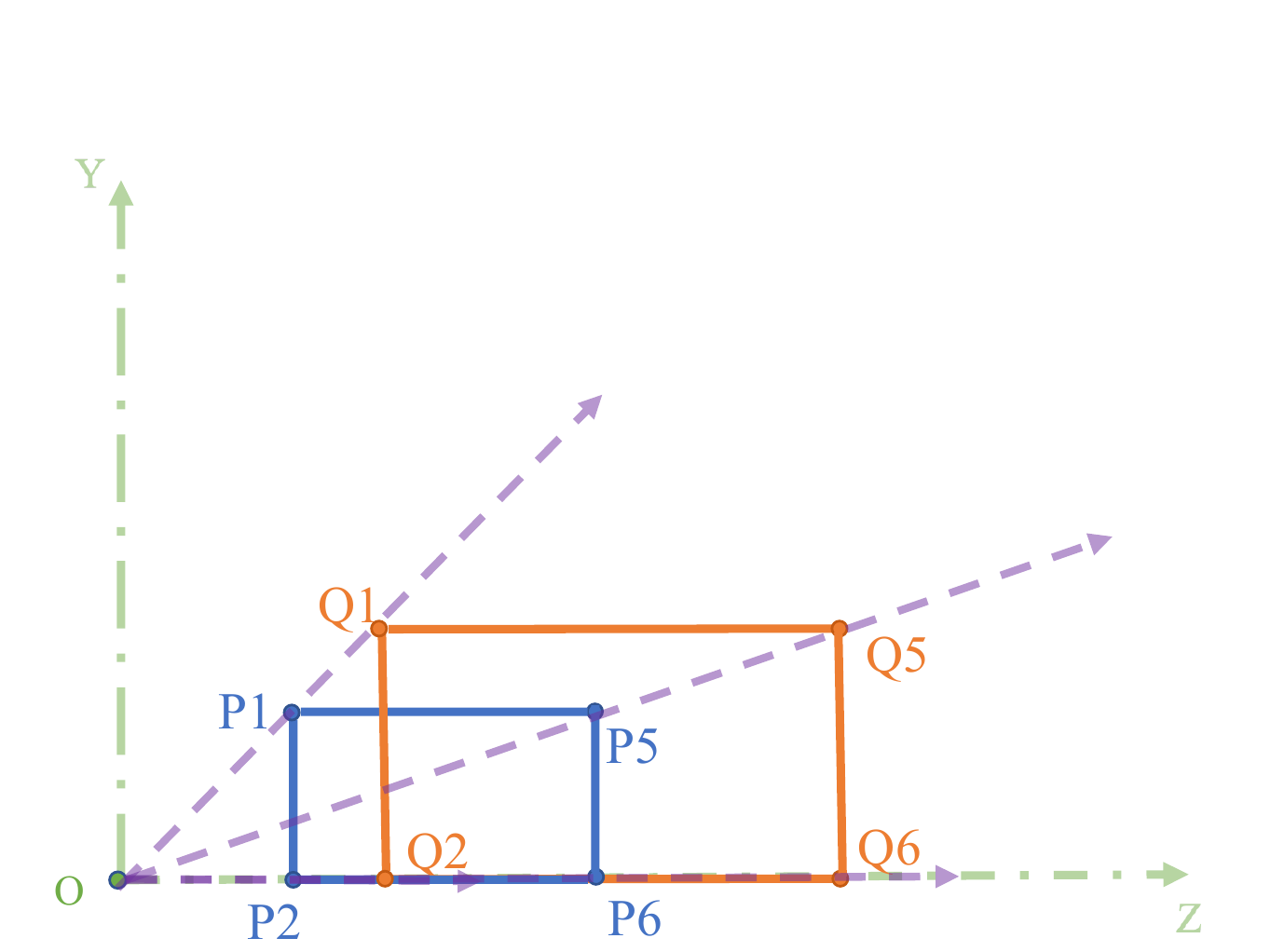}
  \label{fig:side}}
  \hfill
  \subfloat[\scriptsize{BEV with $\text{yaw}\not=0$.}]{\includegraphics[width=0.47\linewidth]{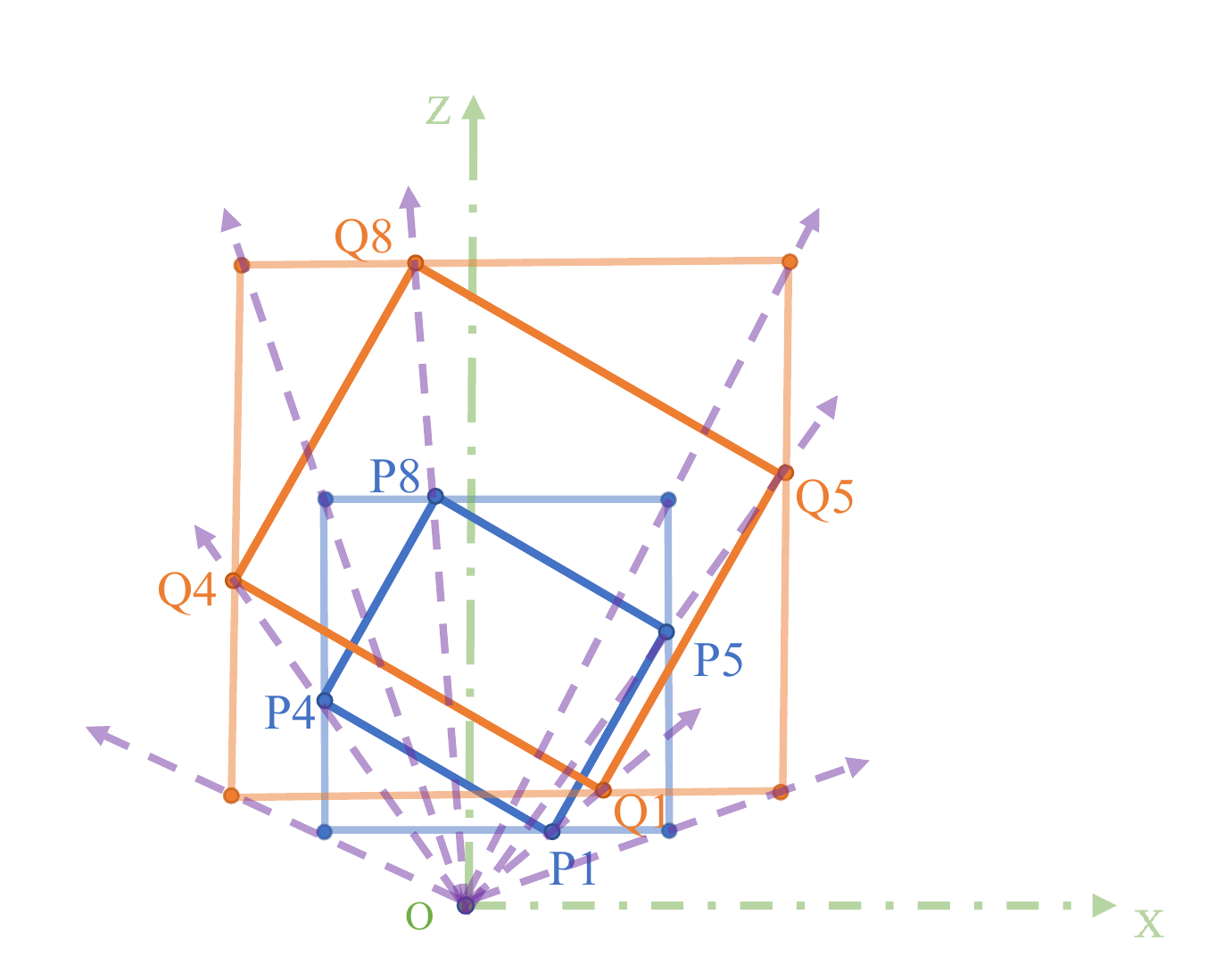}
  \label{fig:alpha}}
  \caption{Different views of object $P$ and $Q$ with the same 2D bounding box.}
  \label{fig:short}
\end{figure}

Moreover, in Figure~\ref{fig:short}, we give an intuitive explanation in three different views: 3D view, bird's eye view (BEV), and lateral view. From Figure~\ref{fig:BEV}, we can see that the ratio of the widths (lengths) equals the ratio of depths. From Figure~\ref{fig:side}, we know that the ratio of heights (lengths) equals the ratio of depths. More generally, when $\text{yaw}\not=0$, we can get the same conclusion using bounding boxes, illustrated in Figure~\ref{fig:alpha}.

In order to obtain the accurate value of depth, an intuitive solution is to estimate object dimensions both in 2D image and 3D space, then recover the depth according to geometry projection. However, the error caused by dimension estimation will amplify the depth estimation error, and it is non-trivial to predict object dimensions precisely. 

Assume the error in dimension estimation is at the centimeter level, then the depth error is $\pm 0.01\times \text{depth}$.
Taking car $P$ and car $Q$ in Figure~\ref{Fig: camera} as an example, assume the dimension scale factor between object $Q$ and object $P$ is $1.02$. For objects in $100$-meter away, as the typical height of cars is $1.53$-meter (averaged value in KITTI), $0.03$-meter dimension errors ($1.53\times (1.02-1) \approx0.03$) can cause $2$-meter depth errors ($100\times (1.02-1)=2$). It will significantly decrease IoU values between predictions and ground truths, which increases training difficulty and instability. It indicates that only using the dimension to resolve the depth is also infeasible.

\subsection{OBMO Module}
\label{Sec: OBMO}
The depth ambiguity causes that objects with different depths can appear very similar visual clues on the RGB image. For monocular-based methods, they have to distinguish depth from such non-discriminative features. This characteristic significantly affects the training stability. 
Therefore, we propose a module named OBMO to resolve the intractable depth ambiguity problem.

OBMO aims to let the network know that objects with different positions in 3D space may have similar bounding boxes and visual features in the 2D image. After looking at multiple reasonable pseudo labels, the network can give more general answers.
Similar to label smoothing~\cite{labelsmooth}, which strengthens the network generalization ability by changing the one-hot encoding to a soft encoding that carries more information. 
Specifically, OBMO is a plug-and-play module capable of being applied during training to any monocular 3D detector.\looseness=-1

\begin{figure}[tb]
\centering
\includegraphics[width=\linewidth]{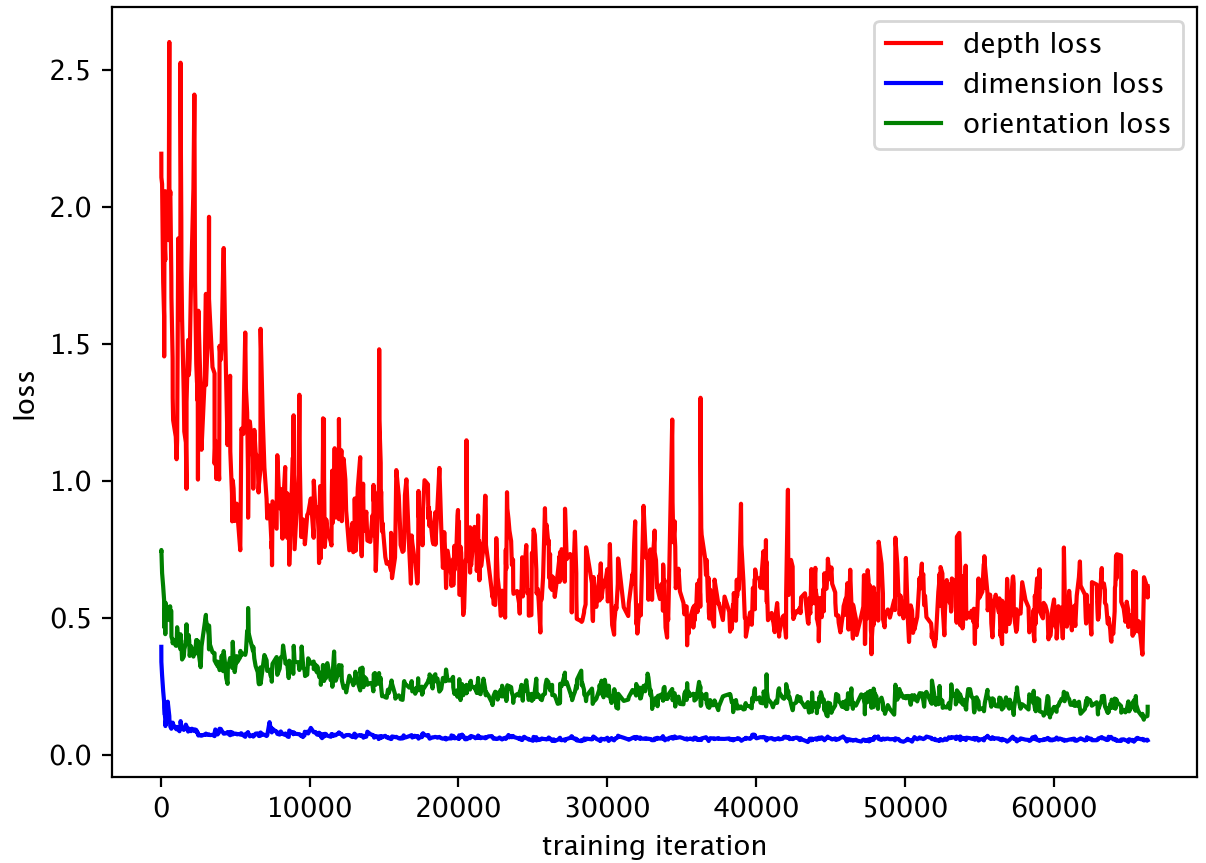}
\caption{The loss based on PatchNet~\cite{PatchNet}. We can see that both predictions of dimension and orientation are stable.}
\label{Fig: dimloss}
\end{figure}

\begin{figure*}[t]
\centering
   \includegraphics[width=0.65\linewidth]{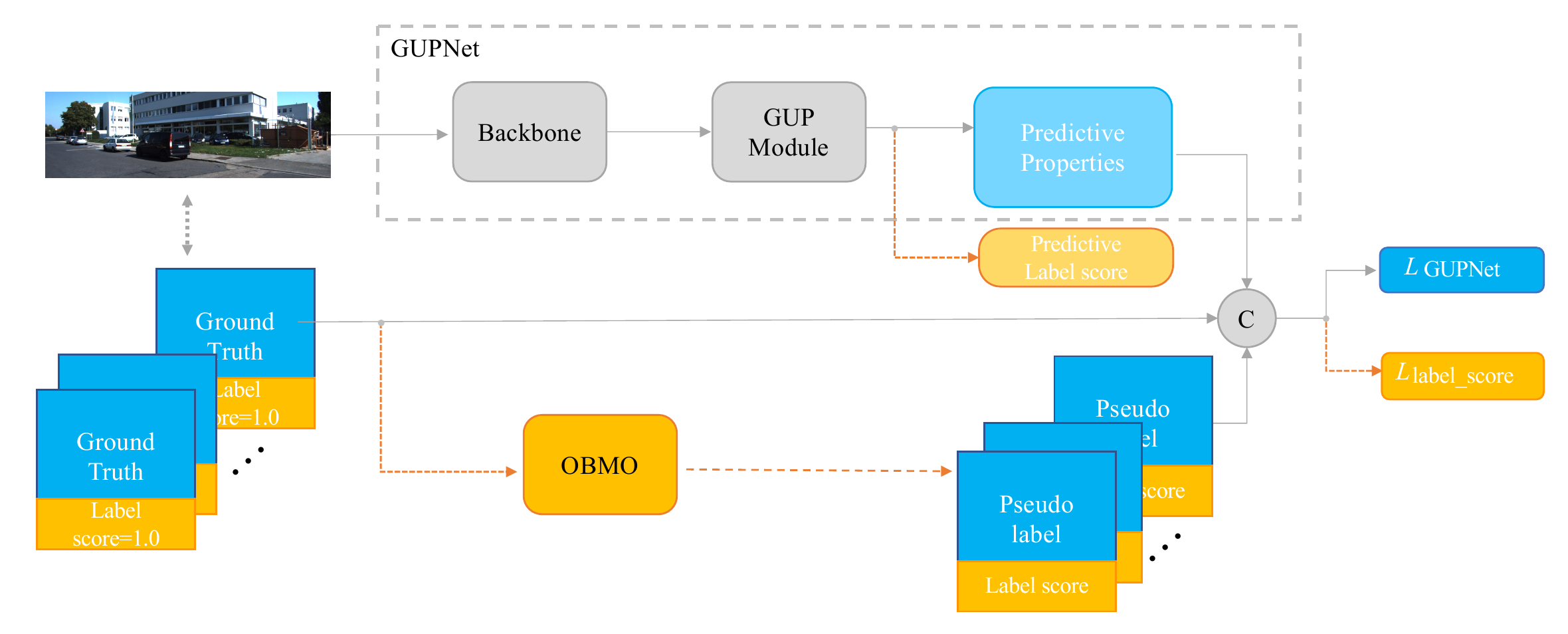}
   \includegraphics[width=0.3\linewidth]{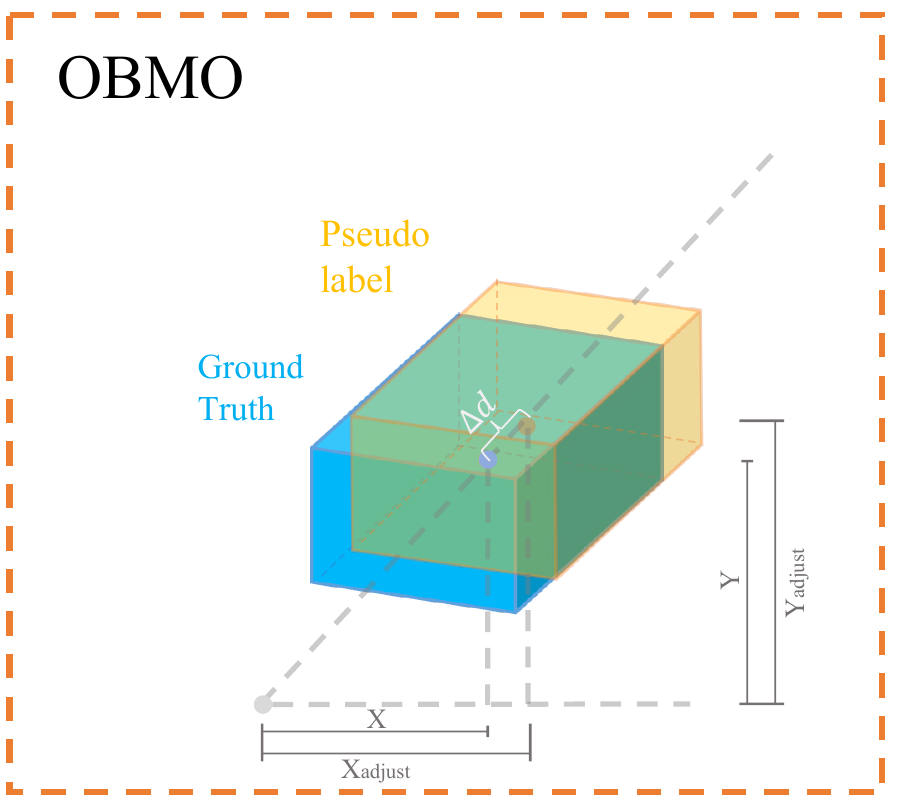}
\caption{The architecture of OBMO with label scoring strategies embedded on GUPNet. The differences are marked in \textcolor[RGB]{255,165,0}{orange}. \textcircled{c} means ``compare''. The OBMO module is used to produce a set of pseudo labels and adds an extra attribute to measure their quality. The Label Score branch is inserted into GUPNet, parallel with 3D prediction branches. Moreover, the OBMO module only works in the training stage.}
\label{Fig: network}
\end{figure*}

To mitigate the adverse impacts of the depth ambiguity issue, we add some pseudo labels along with the viewing frustum within a reasonable range, as shown in Figure~\ref{Fig: network}.
This design improves the generalization ability of the network, as pseudo labels in a larger space remove the strict limitation of original hard labels.
Specifically, we first calculate the X-Z ratio and Y-Z ratio for each object as defined in Equation~\ref{Eq: simple}. Then, we disturb the depth by a set of small offsets for each ground truth $(\text{class}, X, Y, Z, H, W, L,\text{yaw})$.
The depth offsets $\Delta_z$ are determined by the dimension error tolerance and its depth $Z$.
Taking the ``Car'' as an example, we consider the values of $\Delta_z$ from the set $\{-8\%, -4\%, +4\%, +8\%\}$, resulting in $Z_\textrm{adjust}=\Delta_z\cdot Z$. Then, adjust $X$ and $Y$ based on the X-Z ratio and Y-Z ratio, respectively. 
Given that the prediction of the dimensions $(H, W, L)$ is relatively precise and consistent, as depicted in Figure~\ref{Fig: dimloss}, and considering that dimensions are inherent attributes of an object, we preserve their initial values without any alterations, directing our efforts solely towards enhancing the forecast of the 3D location.
To validate our design approach, we conduct ablation studies to justify our design, as presented in Ablation Study Table~\ref{Tab: dim}.
Therefore, we get a new pseudo label $(\textrm{class}, X_\textrm{adjust}, Y_\textrm{adjust}, Z_\textrm{adjust}, H, W, L, \text{yaw})$.
To facilitate learning, we provide ground truths and pseudo labels as supervised signals to the network. It allows the model to incorporate and benefit from the knowledge encoded in the ground truths and the generated pseudo labels during training.

\subsection{Two Label Scoring Strategies}
\label{sec: score}
However, the added pseudo labels are not unlimited. If the pseudo label is too far from the corresponding ground truth, then the transformation of dimensions is too heavy. Moreover, for each category of object, its dimension is limited. Therefore, to make pseudo labels reasonable, we should add constraints on depth offsets or distinguish unreasonable pseudo labels so that irrational pseudo labels will not affect training. Consequently, we design two kinds of label scores, which are used to represent the quality of pseudo labels. One is IoU Label Scores, and the other is Linear Label Scores. Both measure the similarity between pseudo labels and ground truths so they can be obtained before training.

\subsubsection{IoU Label Scores}
Since Intersection over Union (IoU) is a good measurement of how similar two bounding boxes are, we use it as the quality score. The higher the IoU value, the more significant the pseudo label. If two objects do not intersect, the 3D IoU is 0, but the 2D project IoU may not. It is common in categories with smaller lengths, such as Pedestrian. Therefore, instead of using 3D IoU, we use the IoU value of 2D project bounding boxes, defined as follows:
\begin{equation}
\textrm{IoU Label Score} = IoU\left(B_{\textrm{gt}}, B_{\textrm{pseudo}}\right),
\label{Eq: range_IOU}
\end{equation}
$B_{gt}$ refers to the original ground-truth 2D project bounding box, and $B_{pseudo}$ is the projected project bounding 2D box of the added pseudo 3D box label.

% \noindent\textbf{Linear Label Scores}.
\subsubsection{Linear Label Scores}
Furthermore, we introduce another simple yet effective scoring strategy: the Linear Label Score.
It only cares about the offset of depth, and we use a simple linear function, as Equation~\ref{Eq: range} shows,

\begin{equation}
\textrm{Linear Label Score} = 1 - \frac{\left(\lvert\Delta_z\cdot Z\rvert\right)}{c},
\label{Eq: range}
\end{equation}

\noindent where $c$ is a hyper-parameter, and we use it to balance the impacts of pseudo labels. The larger $c$ is, the more enormous impacts pseudo labels have on the training stage. Thus there is a trade-off in the choice of $c$. In our experiments, we choose $c=4$ which empirically makes the score range in $[0,1]$. This scoring strategy intuitively reflects the quality of pseudo labels. For pseudo objects too far away, Linear Label Scores are less than 0 and filter them out. 

For ground truths, the quality scores under both scoring strategies are set to $1.0$. In section~\ref{Sec: ablation} of the ablation study, we find these two label scoring strategies have similar performance, which means that OBMO is robust to the label scoring strategies.

The quality score estimation branch is an auxiliary network that adopts the same structure as the other parallel regression heads. We use L1 loss between the ground truth Label Score and predicted Label Score, as follows:
 
\begin{equation}
\mathcal{L}_\textrm{Label Score} = \lvert\textrm{Label Score}_{\textrm{pred}} - \textrm{Label Score}_{\textrm{gt}}\rvert.
\label{Eq: loss}
\end{equation}

So, the total objective function is:
\begin{equation}
\mathcal{L}_\textrm{total} = \mathcal{L}_\textrm{Baseline} + \lambda\mathcal{L}_\textrm{Label Score},
\label{Eq: loss_total}
\end{equation}
\noindent where $\lambda$ is a trade-off between our Label Score Loss and the losses designed in the original method. If the baseline monocular detector is GUPNet, then $\mathcal{L}_\textrm{Baseline}$ is the hierarchical task loss of 2D detection (including heatmap, 2D offset and 2D size), 3D heads (containing angle, 3D offset and 3D size) and depth inference.

The whole process of the OBMO with label scoring strategies embedded on GUPNet is shown in Figure~\ref{Fig: network}, which can conclude as \textit{adding reasonable pseudo labels} and \textit{adding a parallel Label Score branch}. 
%-------------------------------------------------------------------------
\section{Experiments}
%-------------------------------------------------------------------------
\subsection{Implementation Details}
We adopt published codes from each baseline\footnote{The codes we referenced are: \url{https://github.com/xinzhuma/patchnet} (PatchNet, Pseudo-LiDAR), \url{https://github.com/Owen-Liuyuxuan/visualDet3D} (Ground-aware, RTM3D), and \url{https://github.com/SuperMHP/GUPNet} (GUPNet).}: PatchNet~\cite{PatchNet}, Pseudo-LiDAR~\cite{Pseudo-LiDAR}, Ground-aware~\cite{Ground-aware}, RTM3D~\cite{RTM3D}, and GUPNet~\cite{GUPNet}.  
We use the same configuration described in their papers or projects. Take GUPNet as an example; we use DLA-34 as the backbone, train the model with the batch size of 32 for 140 epochs and adopt the initial learning rate $1.25e^{-3}$ with decay in the 90-th and the 120-th epoch. We train all models on Nvidia GTX 1080Ti GPUs with 11 GB memory.

Moreover, we set $\Delta_Z=\{-8\%, -4\%, +4\%, +8\%\}$ for all detectors and report the better one between IoU Label Score and Linear Label Score.
For the monocular 3D detectors like PatchNet, which regard the scores of 2D bounding boxes as absolute confidence of objects directly, we employ the 2D-3D confidence mechanism from~\cite{OCM3D} to make the scores better describe the 3D predictions. 

%-------------------------------------------------------------------------
%-------------------------------------------------------------------------
\begin{table}[tb]
\centering
\caption{Comparisons on KITTI test set. For easy to compare, we sort them according to their 3D performance on the moderate level of the test set (same as the KITTI leaderboard). We use \textcolor[RGB]{255,0,0}{red} for the highest ones and \textcolor[RGB]{0,0,255}{blue} for the second-highest ones.}
\begin{tabular}{@{}lcccccc@{}}
\toprule
\multirow{2}{*}{Methods} & \multicolumn{3}{c}{$AP_{BEV}$(\%)} & \multicolumn{3}{c}{$AP_{3D}$(\%)} \\
& \text{Easy} & \text{Mod.} & \text{Hard} & \text{Easy} & \text{Mod.} & \text{Hard}\\
\midrule
M3D-RPN~\cite{M3D-RPN} & 21.02 & 13.67 & 10.23 & 14.76 & 9.71 & 7.42\\
SMOKE~\cite{SMOKE} & 20.83 & 14.49 & 12.75 & 14.03 & 9.76 & 7.84\\
RTM3D~\cite{RTM3D} & 19.17 & 14.20 & 11.99  & 14.41 & 10.34 & 8.77\\
PatchNet~\cite{PatchNet} & 22.97 & 16.86 & 14.97 & 15.68 & 11.12 & 10.17\\
KM3D~\cite{KM3D} & 23.44 & 16.20 & 14.47 & 16.73 & 11.45 & 9.92\\
D4LCN~\cite{D4LCN} & 22.51 & 16.02 & 12.55 & 16.65 & 11.72 & 9.51\\
Monodle~\cite{Monodle} & 24.79 & 18.89 & 16.00 & 17.23 & 12.26 & 10.29\\
MonoRUn~\cite{MonoRUn} & 27.94 & 17.34 & 15.24 & 19.65 & 12.30 & 10.58\\
GrooMeD-NMS~\cite{GrooMeD-NMS} & 26.19 & 18.27 & 14.05 & 18.10 & 12.32 & 9.65\\
Ground-aware\cite{Ground-aware} & \textcolor[RGB]{0,0,255}{29.81} & 17.98 & 13.08 & \textcolor[RGB]{0,0,255}{21.65} & 13.25 & 9.91\\
CaDDN~\cite{CaDDN} & 27.94 & 18.91 & 17.19 & 19.17 & 13.41 & 11.46\\
MonoEF~\cite{MonoEF} & 29.03 & 17.26 & \textcolor[RGB]{255,0,0}{19.70} & 21.29 & 13.87 & 11.71\\
MonoFlex~\cite{MonoFlex} & 28.23 & \textcolor[RGB]{0,0,255}{19.75} & 16.89 & 19.94 & 13.89 & \textcolor[RGB]{0,0,255}{12.07}\\
GUPNet~\cite{GUPNet} & - & - & - & 20.11	& \textcolor[RGB]{0,0,255}{14.20}	& 11.77\\
\hline
GUPNet (+ OBMO) & \textcolor[RGB]{255,0,0}{30.81} & \textcolor[RGB]{255,0,0}{21.41} & \textcolor[RGB]{0,0,255}{18.37} & \textcolor[RGB]{255,0,0}{22.71} & \textcolor[RGB]{255,0,0}{15.70} & \textcolor[RGB]{255,0,0}{13.23}\\
Improvements & - & - & - & +2.6 & +1.50 & +1.46\\
\bottomrule
\end{tabular}
\label{Tab: compare}
\end{table}
\subsection{Dataset and Metrics}
We conduct experiments on KITTI~\cite{KITTI} and Waymo~\cite{Waymo} benchmarks. 

\subsubsection{KITTI}
KITTI is the widely employed dataset for monocular 3D object detection. 
It provides 7481 images for training and 7518 images for testing. 
All the scenes are pictured around Karlsruhe, Germany in clear weather and day time.
To make fair comparisons, we follow previous works~\cite{split,Ground-aware,GUPNet} to split the training images into train set (3712 images) and val set (3769 images). 
All experiments are performed under this dataset split.
Furthermore, the detection results are evaluated under three levels of difficulty: easy, moderate and hard, which are defined according to the height of the 2D bounding box, occlusion, and truncation. 
We conduct experiments under two core evaluations: the average precision of 3D bounding boxes $AP_{3D}$ and the average precision of objects in Bird's Eye View $AP_{BEV}$. 
For the metric, we employ the recently suggested metric $AP_{BEV|R_{40}}$ and $AP_{3D|R_{40}}$ by KITTI benchmark~\cite{KITTI}.
Following common practice~\cite{PatchNet,RTM3D,GUPNet}, we evaluate the results on the Car category under IoU threshold 0.7.

\subsubsection{Waymo}
The Waymo dataset is a recently released large dataset for autonomous driving research.
It consists of 798 training sequences and 202 validation sequences. 
The scenes are pictured in Phoenix, Mountain View, and San Francisco under multiple kinds of weathers and at multiple times of a day. 
Different from KITTI, it provides 3D box labels in the $360$-degree field of view, while we only use the front view for the task of monocular 3D object detection.
We use the same data processing strategy proposed in CaDDN~\cite{CaDDN}. Specifically, we sample every third frame from the training sequences to form our training set due to the large dataset size and high frame rate.
We adopt the officially released evaluation to calculate the mean average precision (mAP) and the mean average precision weighted by heading (mAPH).
The evaluation is separated by difficulty setting (LEVEL 1, LEVEL 2) and distance to the sensor ($0 - 30$ m, $30 - 50$ m, and $50$ m $ - \infty$). We evaluate the Car category with IoU criteria of $0.7$ and $0.5$.\looseness=-1

\begin{figure}[t]
\centering
\includegraphics[width=\linewidth]{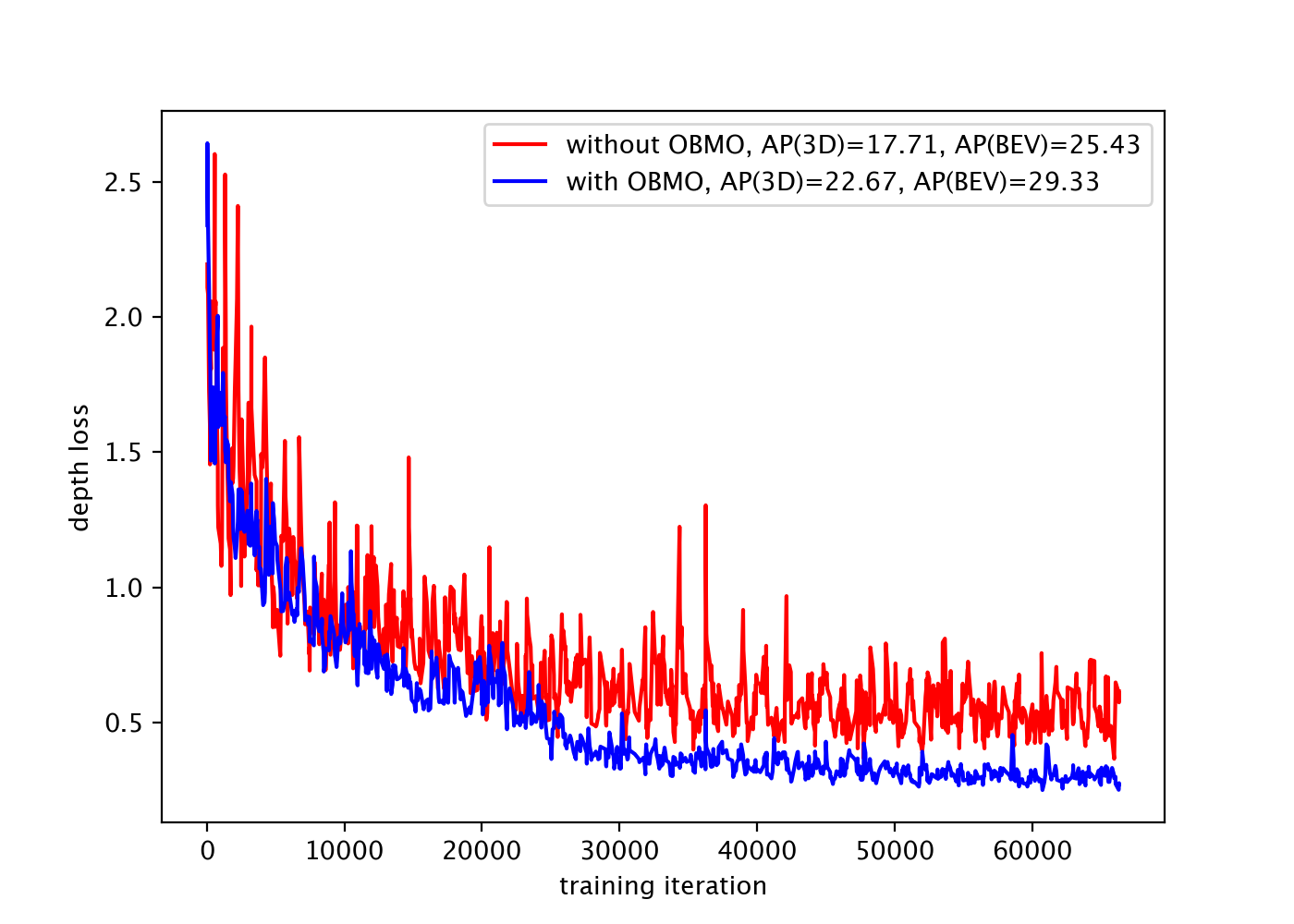}
\caption{The depth loss with/without the proposed module (OBMO) based on PatchNet. We can see that OBMO can stabilize depth training from the more stable loss curve.}
\label{Fig: depth}
\end{figure}

\subsection{Quantitative Results}
In Table~\ref{Tab: compare}, we conduct a comprehensive comparison between our proposed method and existing state-of-the-art methods on the test sets of KITTI benchmark for Car. Without bells and whistles, our method outperforms all prior methods under $AP_{BEV|R_{40}}$ and $AP_{3D|R_{40}}$ including those with extra information. For $AP_{3D|R_{40}}$, our method is $22.71\%/15.70\%/13.23\%$, which is much higher than the baseline GUPNet on three levels of difficulty. The performance improvement is even more significant at the easy level. We suspect this is because, for foreign vehicles, the tiny depth shift needs a significant dimension transformation compared to near cars. Indeed, there is also less visual information in distant objects.\looseness=-1

\begin{table*}[tb]
\centering
\caption{Performance of using our OBMO method, including OBMO module and quality scores, on different SOTA monocular detectors. All methods are evaluated on KITTI val set with metric $AP|_{R_{40}}$. We report both the results of using the IoU Label Score and Linear Label Score. And we only compare \colorbox{lightgray}{the best one} of $AP_{BEV}$ Mod. with the baseline.}
\begin{tabular}{@{}lccccccccc@{}}
\toprule
\multirow{2}{*}{Methods} & \multicolumn{3}{c}{$AP_{2D}$(\%)} & \multicolumn{3}{c}{$AP_{BEV}$(\%)} & \multicolumn{3}{c}{$AP_{3D}$(\%)} \\
& \text{Easy} & \text{Mod.} & \text{Hard} & \text{Easy} & \text{Mod.} & \text{Hard} & \text{Easy} & \text{Mod.} & \text{Hard}\\
\midrule
PatchNet~\cite{PatchNet} & 97.05 & 94.00 & 86.32 & 43.97 & 25.43 & 20.73 & 32.56 & 17.71 & 13.98\\
(+ 3D scores~\cite{OCM3D})  & 97.17 & 94.07 & 87.00 & 41.82 & 28.13 & 24.23 & 32.96 & 21.27 & 17.87\\
\rowcolor{lightgray}(+ OBMO-Linear Label Scores) & 97.20 & 94.00 & 86.96 & 42.91 & 29.33 & 24.53 & 34.16 & 22.67 & 18.22\\
(+OBMO-IoU Label Scores) & 97.23 & 93.96 & 86.92 & 43.10 & 29.47 & 24.59 & 33.24 & 22.30 & 18.15\\
Improvements & \bf{+0.15} & 0.00 & \bf{+0.64} & -1.06 & \bf{+3.90} & \bf{+3.80} & \bf{+1.60} & \bf{+4.96} & \bf{+4.24}\\
\hline
PatchNet* & 97.05 & 94.00 & 86.32 & 26.03 & 15.05 & 12.74 & 18.54 & 10.46 & 8.71\\
(+ 3D scores)  & 97.63 & 94.34 & 87.17 & 30.03 & 20.68 & 17.60 & 22.01 & 15.37 & 12.83\\
\rowcolor{lightgray}(+ OBMO-Linear Label Scores) &  97.57 & 94.22 & 87.11 & 32.41 & 22.75 & 19.56 & 24.40 & 16.63 & 14.53\\
(+ OBMO-IoU Label Scores) &  97.41 & 94.12 & 87.03 & 32.72 & 22.58 & 19.24 & 24.95 & 16.60 & 14.20\\
Improvements &  \bf{+0.52} & \bf{+0.22} & \bf{+0.79} &  \bf{+6.38} &  \bf{+7.70} &  \bf{+6.82} &  \bf{+5.86} &  \bf{+6.17} &  \bf{+5.82}\\
\hline
Pseudo-LiDAR$\dagger$~\cite{Pseudo-LiDAR} & 97.05 & 94.00 & 86.32 & 37.77 & 21.31 & 17.92 & 25.19 & 12.72 & 10.22\\
(+ OBMO-Linear Label Scores) &  95.18 & 92.68 & 86.05 & 36.76 & 24.72 & 21.01 & 24.39 & 16.07 & 13.32\\
\rowcolor{lightgray}(+ OBMO-IoU Label Scores) & 96.27 & 92.90 & 86.14 & 39.55 & 25.40 & 20.85 & 25.78 & 16.03 & 13.13\\
Improvements & -0.78 & -1.10 & -0.18 & \bf{+1.78} & \bf{+4.09} & \bf{+2.93} & \bf{+0.59} & \bf{+3.31} & \bf{+2.91}\\
\hline
RTM3D$\dagger$ & 97.02 & 91.49 & 83.98 & 21.27 & 15.84 & 13.63 & 15.05 & 11.42 & 9.66\\ 
\rowcolor{lightgray}(+ OBMO-Linear Label Scores) & 97.03 & 91.44 & 83.93 & 23.25 & 17.60 & 14.93 & 16.50 & 12.74 & 10.59\\
(+ OBMO-IoU Label Scores) & 97.03 & 91.45 & 83.93 & 22.99 & 17.52 & 14.83 & 16.29 & 12.75 & 10.54\\
Improvements & \bf{+0.01} & -0.05 & -0.05 & \bf{+1.98} & \bf{+1.76} & \bf{+1.30} & \bf{+1.45} & \bf{+1.32} & \bf{+0.93}\\
\hline
Ground-aware~\cite{Ground-aware} & - & - & - & 28.95 & 20.11 & 15.51 & 22.80 & 15.41 & 11.43 \\
\rowcolor{lightgray}(+ OBMO-Linear Label Scores)  & 96.79 & 84.04 & 64.26 & 31.22 & 21.96 & 16.85 & 23.48 & 16.59 & 12.39\\
(+ OBMO-IoU Label Scores) & 96.83 & 81.72 & 61.90 & 29.58 & 21.75 & 16.73 & 22.64 & 16.40 & 12.22\\
Improvements & - & - & - & \bf{+2.27} & \bf{+1.85} & \bf{+1.34} & \bf{+0.68} & \bf{+1.18} & \bf{+0.96}\\
\hline
GUPNet~\cite{GUPNet} & - & - & - & 31.07 & 22.94 & 19.75 & 22.76 & 16.46 & 13.72 \\
(+ OBMO-Linear Label Scores) & 96.67 & 88.67 & 78.85 & 33.09 & 23.63 & 20.42 & 24.65 & 17.80 & 15.15\\
\rowcolor{lightgray}(+ OBMO-IoU Label Scores) & 96.58 & 88.64 & 78.92 & 32.20 & 23.88 & 20.67 & 24.48 & 17.94 & 15.26\\
Improvements & - & - & - & \bf{+1.13} & \bf{+0.94} & \bf{+0.92} & \bf{+1.72} & \bf{+1.48} & \bf{+1.54}\\
\bottomrule
\end{tabular}
\label{Tab: general}
\end{table*}

We further show the efficiency of our module OBMO embedded in other different SOTA monocular detectors in Table~\ref{Tab: general}. Because of the different train-val split, \text{PatchNet*} is retrained by its public code with a unified split~\cite{split}. As for RTM3D and Pseudo-LiDAR, which only report results on $AP|_{R_{11}}$ in their paper, we evaluated them on $AP|_{R_{40}}$ by their public models (use $\dagger$ to represent). 
The improvements indicate that OBMO can be applied both in direct regression-based and depth-aware methods. 
The results show that the improvement in depth-aware methods is more remarkable than in direct regression-based methods. 
Specifically, for PatchNet, we improve the $AP_{BEV}/AP_{3D}$ from $25.43\%/17.71\%$ to $29.33\%/22.67\%$ under the moderate setting. We think that OBMO might mitigate the influence of the worse monocular depth estimation to a certain extent.
For direct regression-based methods such as RTM3D, the original detector is boosted by $1.82\%/2.14\%$ in $AP_{BEV}/AP_{3D}$ under the moderate setting. 
Such significant improvements demonstrate the effectiveness and robustness of our method.
We also present the 2D mAP in Table~\ref{Tab: general}. The 2D performances with and without OBMO are similar because the reasonable pseudo labels produced by the OBMO module are along the viewing frustum.
Note the 2D detectors used in PatchNet and Pseudo-LiDAR are both Faster-RCNN, so their 2D mAPs are the same.

\begin{figure*}[tb]
\centering
   \includegraphics[width=0.45\linewidth]{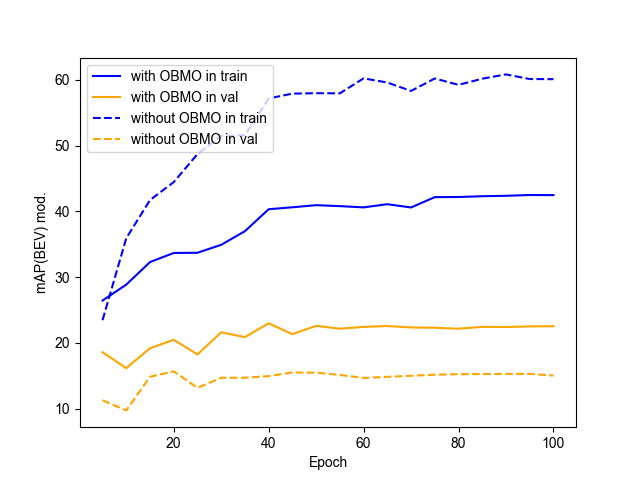}
   \includegraphics[width=0.45\linewidth]{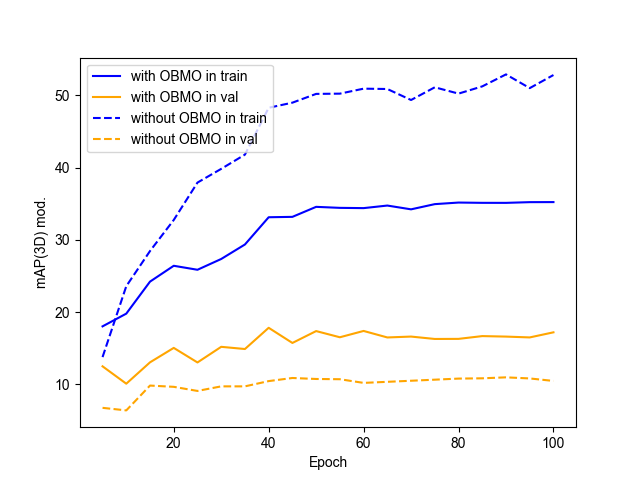}
\caption{The mAP (3D/BEV) with/without OBMO based on PatchNet in training and validation set. We can see that OBMO can overcome overfitting to a certain extent.\looseness=-1}
\label{Fig: mAP}
\end{figure*}

We further investigate the depth loss curve and the mAP curves after adding the OBMO module in training.
As in Figure~\ref{Fig: depth}, we can easily see that the method employing OBMO has a smoother learning curve.
By contrast, the curve of the original detector is unstable and contains many strong oscillations.
It indicates that OBMO endows the network to steadily learn depth, stabilizing the overall learning process and thus bringing apparent improvements.
As for the mAP curves shown in Figure~\ref{Fig: mAP}, it is not difficult to find that without the OBMO module, the mAP of the training set is still rising, while there is almost no fluctuation in the mAP of the validation set or even a slight decline. 
It can be illustrated that OBMO overcomes overfitting to a certain extent.

\subsection{Ablation Studies}
\label{Sec: ablation}
We take PatchNet\text{*} as our baseline detector in the ablation study to save training time. 
By default, we set depth offset $\Delta_z$ to $\{-8\%, -4\%, +4\%, +8\%\}$ and use Linear Label Score. 
\text{X-Z} ratio and \text{Y-Z} ratio are both regarded as constraints. 

\begin{table}[tb]
\centering
\caption{Ablation study on each component in OBMO. The 3D score~\cite{OCM3D} combines the 2D score with 3D information to represent the score of an object.}
\resizebox{\columnwidth}{!}{
\begin{tabular}{@{}cccccccccc@{}}
\toprule
\text{3D} & \multirow{2}{*}{OBMO} & \text{IoU Label} & \text{Linear Label} & \multicolumn{3}{c}{$AP_{BEV}$(\%)} & \multicolumn{3}{c}{$AP_{3D}$(\%)} \\
\text{scores} &  & Scores & Scores & Easy & Mod. &  Hard& Easy & Mod. &  Hard\\
 \midrule
 & & & & 26.03 & 15.05 & 12.74 & 18.54 & 10.46 & 8.71\\
\checkmark & & & & 30.03 & 20.68 & 17.60 & 22.01 & 15.37 & 12.83\\
\checkmark & \checkmark & & & 31.13 & 22.14 & 19.02 & 23.87 & 16.41 & 14.25\\
\checkmark & \checkmark & \checkmark & &  \bf{32.72} & 22.58 & 19.24 & \bf{24.95} & 16.60 & 14.20\\
\checkmark & \checkmark & & \checkmark &  32.41 & \bf{22.75} & \bf{19.56} & 24.40 & \bf{16.63} & \bf{14.53}\\
\bottomrule
\end{tabular}}
\label{Tab: submodule}
\end{table}

\begin{table}[t]
\centering
\caption{Ablation study on different constraints.}
\resizebox{.98\columnwidth}{!}{
\begin{tabular}{@{}cccccccc@{}}
\toprule
\text{under} &  under & \multicolumn{3}{c}{$AP_{BEV}$(\%)} & \multicolumn{3}{c}{$AP_{3D}$(\%)} \\
X-Z ratio & Y-Z ratio & \text{ Easy } & \text{ Mod. } & \text{ Hard } & \text{ Easy } & \text{ Mod. } & \text{ Hard }\\
\midrule
 & & 30.03 & 20.68 & 17.60 & 22.01 & 15.37 & 12.83\\
\checkmark & & 31.12 & 22.30 & 19.18 & 23.36 & 16.36 & 14.09\\
 & \checkmark & 28.73 & 20.44 & 17.34 & 21.31 & 15.28 & 12.64\\
\checkmark & \checkmark& \bf{32.41} & \bf{22.75} & \bf{19.56} & \bf{24.40} & \bf{16.63} & \bf{14.53}\\
\bottomrule
\end{tabular}}
\label{Tab: ratio}
\end{table}

\noindent\textbf{Validity of Each Component}. To study the impact brought by each component of OBMO, we investigate them through extra experiments, as shown in \text{Table~\ref{Tab: submodule}}. 
The results show that each component of the OBMO module is effective. 
As components gradually increase, the final accuracy also increases accordingly.
We can see that the initial performance (AP) is boosted from 
$15.05\%/10.46\%$ to $22.75\%/16.63\%$
under the moderate setting, which is rather impressive.
Moreover, both IoU Label Score and Linear Label Score for pseudo labels work well, suggesting that the proposed soft pseudo label strategy is robust since it is not sensitive for specifically designed manners.

\noindent\textbf{Different Constraints}. Also, we investigate the impact brought by different constraints, namely, the X-Z ratio and Y-Z ratio. If we do not use the X-Z ratio or Y-Z ratio as constraints, X or Y will not change in our pseudo labels.
The results are reported in \text{Table~\ref{Tab: ratio}}. Ultimately, we achieve the best performance by using them in combination.

\begin{table}[tb]
\centering
\caption{Ablation study on the depth offset under the same number of pseudo labels.}
\begin{tabular}{@{}ccccccc@{}}
\toprule
\multirow{2}{*}{\text{  Offset(\%) }} & \multicolumn{3}{c}{$AP_{BEV}$(\%)} & \multicolumn{3}{c}{$AP_{3D}$(\%)} \\
& \text{Easy} & \text{Mod.} & \text{Hard} & \text{Easy} & \text{Mod.} & \text{Hard}\\
\midrule
0 & 30.03 & 20.68 & 17.60 & 22.01 & 15.37 & 12.83\\
\hline
2 & 30.78 & 22.09 & 18.84 & 23.78 & 16.45 & 14.04\\
4 & \bf{32.41} & \bf{22.75} & \bf{19.56} & \bf{24.40} & \bf{16.63} & \bf{14.53}\\
6 & 29.35 & 21.85 & 18.72 & 22.11 & 15.98 & 13.86\\
8 & 29.97 & 21.82 & 18.78 & 22.67 & 15.98 & 13.28\\
\bottomrule
\end{tabular}
\label{Tab: offset}
\end{table}

\begin{table}[tb]
\centering
\caption{Ablation study on the number of pseudo labels in the same depth offset.} 
\begin{tabular}{@{}ccccccc@{}}
\toprule
\text{ the number of } & \multicolumn{3}{c}{$AP_{BEV}$(\%)} & \multicolumn{3}{c}{$AP_{3D}$(\%)} \\
\text{pseudo labels} & \text{Easy} & \text{Mod.} & \text{Hard} & \text{Easy} & \text{Mod.} & \text{Hard}\\
\midrule
0 & 30.03 & 20.68 & 17.60 & 22.01 & 15.37 & 12.83\\
\hline
2 & 31.17 & 21.96 & 18.83 & 24.43 & 16.43 & 13.64\\
4 & \bf{32.41} & \bf{22.75} & \bf{19.56} & \bf{24.40} & \bf{16.63} & \bf{14.53}\\
6 & 30.41 & 22.09 & 18.99 & 22.98 & 16.02 & 13.29\\
8 & 31.14 & 22.32 & 19.04 & 23.65 & 16.41 & 14.21\\
\bottomrule
\end{tabular}
\label{Tab: number2}
\end{table}

% \iffalse
\begin{table}[tb]
\centering
\caption{Ablation study on the number of pseudo labels in the same depth range.}
\begin{tabular}{@{}ccccccc@{}}
\toprule
\text{ the number of } & \multicolumn{3}{c}{$AP_{BEV}$(\%)} & \multicolumn{3}{c}{$AP_{3D}$(\%)} \\
\text{pseudo labels} & \text{Easy} & \text{Mod.} & \text{Hard} & \text{Easy} & \text{Mod.} & \text{Hard}\\
\midrule
0 & 30.03 & 20.68 & 17.60 & 22.01 & 15.37 & 12.83\\
\hline
2 & 29.17 & 21.37 & 18.49 & 21.01 & 15.35 & 12.78\\
4 & \bf{32.41} & \bf{22.75} & \bf{19.56} & \bf{24.40} & \bf{16.63} & \bf{14.53}\\
6 & 30.69 & 22.16 & 18.98 & 23.64 & 16.40 & 13.52\\
8 & 31.82 & 22.66 & 19.23 & 23.68 & 16.47 & 14.03\\
\bottomrule
\end{tabular}
\label{Tab: number}
\end{table}
% \fi 

\noindent\textbf{Depth Offsets $\Delta_z$}. We further show the influences of different depth offsets.
In particular, we have to choose a suitable depth offset carefully due to discrete depth values.
However, there is a dilemma in making a choice.
If the depth offset is too small, we have to add multiple pseudo labels in the reasonable depth range, and the computational complexity will increase dramatically.
On the contrary, if the depth offset is too large, we will lose some reasonable pseudo labels, resulting in suboptimal performance.
Therefore, according to statistical dimensions information of the KITTI dataset, we try four base offset values: $\{2\%, 4\%, 6\%, 8\%\}$ with $4$ pseudo labels. Specially, if we choose the base of $2\%$, then $\Delta_z=\{-4\%, -2\%, +2\%, +4\%\}$.
We report the corresponding results in Table~\ref{Tab: offset}.
It shows that the proper depth offset is indeed desired. 

Then we fix the base value of the depth offset to $4\%$, and change the number of pseudo labels. Specially, if we use four pseudo labels, then $\Delta_z=\{-8\%, -4\%, +4\%, +8\%\}$. The results are in Table~\ref{Tab: number2}. Adding six pseudo labels reduces performance compared to adding four pseudo labels. It means if the depth value is outside the reasonable range, the added pseudo labels do not help performance and can even be detrimental to the performance.

Intuitively, if the added pseudo labels are too dense, it will also decrease the performance. 
Therefore, we set the largest value of depth offset to $8\% \cdot Z$, and choose the number of pseudo labels from $\{2, 4, 6, 8\}$. Specially, if we use $4$ pseudo labels, then $\Delta_z=\{-8\%, -4\%, +4\%, +8\%\}$.
The results are shown in Table~\ref{Tab: number}, which verifies the viewpoint.

In conclusion, the best choice of depth offset $\Delta_z$ for Car category in KITTI dataset is $\{-8\%, -4\%, +4\%, +8\%\}$. We also use this setting in the Waymo dataset because the dimension of the Car category is similar.

\begin{table}[bt]
\centering
\caption{Ablation study on implying range. None means we don't use OBMO module.}
\begin{tabular}{@{}ccccccc@{}}
\toprule
\multirow{2}{*}{ level $\geq$* } & \multicolumn{3}{c}{$AP_{BEV}$(\%)} & \multicolumn{3}{c}{$AP_{3D}$(\%)} \\
& \text{Easy} & \text{Mod.} & \text{Hard} & \text{Easy} & \text{Mod.} & \text{Hard}\\
\midrule
None & 30.03 & 20.68 & 17.60 & 22.01 & 15.37 & 12.83\\
\hline
3 & 27.06 & 19.42 & 16.46 & 19.75 & 13.78 & 11.79\\ % ？
2 & 31.11 & 22.58 & 19.56 & 22.17 & 15.29 & 13.98\\
1 & \bf{32.41} & \bf{22.75} & \bf{19.56} & \bf{24.40} & \bf{16.63} & \bf{14.53}\\
\bottomrule
\end{tabular}
\label{Tab: depth}
\end{table}

\begin{table}[tb]
\centering
\caption{Ablation study on the weight of label score branch.}
\begin{tabular}{@{}ccccccc@{}}
\toprule
\multirow{2}{*}{ Lambda } & \multicolumn{3}{c}{$AP_{BEV}$(\%)} & \multicolumn{3}{c}{$AP_{3D}$(\%)} \\
& \text{Easy} & \text{Mod.} & \text{Hard} & \text{Easy} & \text{Mod.} & \text{Hard}\\
\midrule
0 & 31.13 & 22.14 & 19.02 & 23.87 & 16.41 & 14.25\\
0.5 & 31.98 & 22.46 & 19.31 & 23.05 & 16.21 & 13.47\\
1 & \bf{32.41} & \bf{22.75} & \bf{19.56} & \bf{24.40} & \bf{16.63} & \bf{14.53}\\
2 & 30.07 & 21.15 & 18.72 & 23.19 & 16.20 & 13.46\\
\bottomrule
\end{tabular}
\label{Tab: loss}
\end{table}

Additionally, we evaluate the performance of applying OBMO to different difficulty levels of objects in Table~\ref{Tab: depth}. The difficulty level of the object is defined according to the height of the 2D bounding box, occlusion, and truncation values.
We can see that OBMO works well for all levels of objects, which means that the issue of depth ambiguity indeed exists and is widespread.

\noindent\textbf{Weights of Label Score $\lambda$}. We use different loss weights in the label score branch and report the results in Table~\ref{Tab: loss}. The model performs better when the weight of loss is set to $1$.

\begin{table}[t]
\centering
\caption{Ablation study of the influence in changing dimension.}
\begin{tabular}{@{}lcccccc@{}}
\toprule
\multirow{2}{*}{Change} & \multicolumn{3}{c}{$AP_{BEV}$(\%)} & \multicolumn{3}{c}{$AP_{3D}$(\%)} \\
& \text{Easy} & \text{Mod.} & \text{Hard} & \text{Easy} & \text{Mod.} & \text{Hard}\\
\midrule
None & 30.03 & 20.68 & 17.60 & 22.01 & 15.37 & 12.83\\
Dimension & 29.06 & 16.90 & 14.15 & 21.95 & 11.85 & 9.58\\
Position & \textbf{32.41} & \textbf{22.75} & \textbf{19.56} & \textbf{24.40} & \textbf{16.63} & \textbf{14.53}\\
Both & 7.07 & 6.86 & 5.92 & 3.28 & 3.40 & 3.00\\
\bottomrule
\end{tabular}
\label{Tab: dim}
\end{table}

\begin{table}[tb]\centering
\caption{$AP_{3D|40}$ on Pedestrian and Cyclist on KITTI validation set.}
\resizebox{.98\columnwidth}{!}{
\begin{tabular}{@{}lccccccc@{}}
\toprule
\multirow{2}{*}{ Categories } & \multirow{2}{*}{  Methods  } & \multicolumn{3}{c}{$AP_{BEV}$(\%)} & \multicolumn{3}{c}{$AP_{3D}$(\%)} \\
& & \text{Easy} & \text{Mod.} & \text{Hard} & \text{Easy} & \text{Mod.} & \text{Hard}\\
\midrule
\multirow{3}{*}{Pedestrian} & PatchNet & 10.55 & 8.23 & 6.48 & 8.82 & 6.82 & 5.14\\
 & (+ OBMO) & 16.17 & 11.93 & 9.38 & 12.80 & 9.55 & 7.40\\
 & Improvements & \bf{+5.62} & \bf{+3.70} & \bf{+2.89} & \bf{+3.98} & \bf{+2.73} & \bf{+2.25}\\
\hline
\multirow{3}{*}{Cyclist} & PatchNet & 7.47 & 3.64 & 3.03 & 5.83 & 2.87 & 2.60\\
 & (+ OBMO) & 9.30 & 4.72 & 4.45 & 7.81 & 4.06 & 3.60\\
 & Improvements & \bf{+1.83} & \bf{+1.08} & \bf{+1.42} & \bf{+1.98} & \bf{+1.19} & \bf{+1.00}\\
\bottomrule
\end{tabular}}
\label{Tab: PedCyc}
\end{table}

\noindent\textbf{Keep Dimension}. To verify that changing dimension harms the performance, we change both the position and dimension. The results are shown in Table~\ref{Tab: dim}. When we change the dimensions of the object, the performance drops drastically. We can't change the inherent property of objects. Otherwise, the pseudo labels are not reasonable. As for the positions that we modify, they are current state values and can be changed.

\noindent\textbf{Generalization of OBMO}. Furthermore, we verify the generalization of our OBMO method. On the one hand, we test on other categories: Pedestrian and Cyclist. On the other hand, we test on another larger dataset: Waymo.

For the first one, we use the same default configuration as in Car, i.e., four pseudo labels: $\Delta_z=\{-8\%, -4\%, +4\%, +8\%\}$ and IoU Label Score. The results are shown in Table~\ref{Tab: PedCyc}. The IoU threshold we used is $0.5$ for both of them. The improvements are apparent, proving that our method can apply to varied categories.

For the latter one, we take GUPNet~\cite{GUPNet} as our baseline and adopt the metrics with mAP and mAPH under the IoU threshold of $0.7$ and $0.5$, respectively. “Level 1” denotes the evaluation of the bounding boxes that contain more than $5$ lidar points. “Level 2” denotes the evaluation of all bounding boxes. The results prove that our proposed OBMO method achieves consistent improvements in all settings, as shown in Table~\ref{Tab: waymo}.\looseness=-1

%-------------------------------------------------------------------------
\subsection{Qualitative Results}
\begin{figure*}[tbp]
\centering
\includegraphics[width=\textwidth]{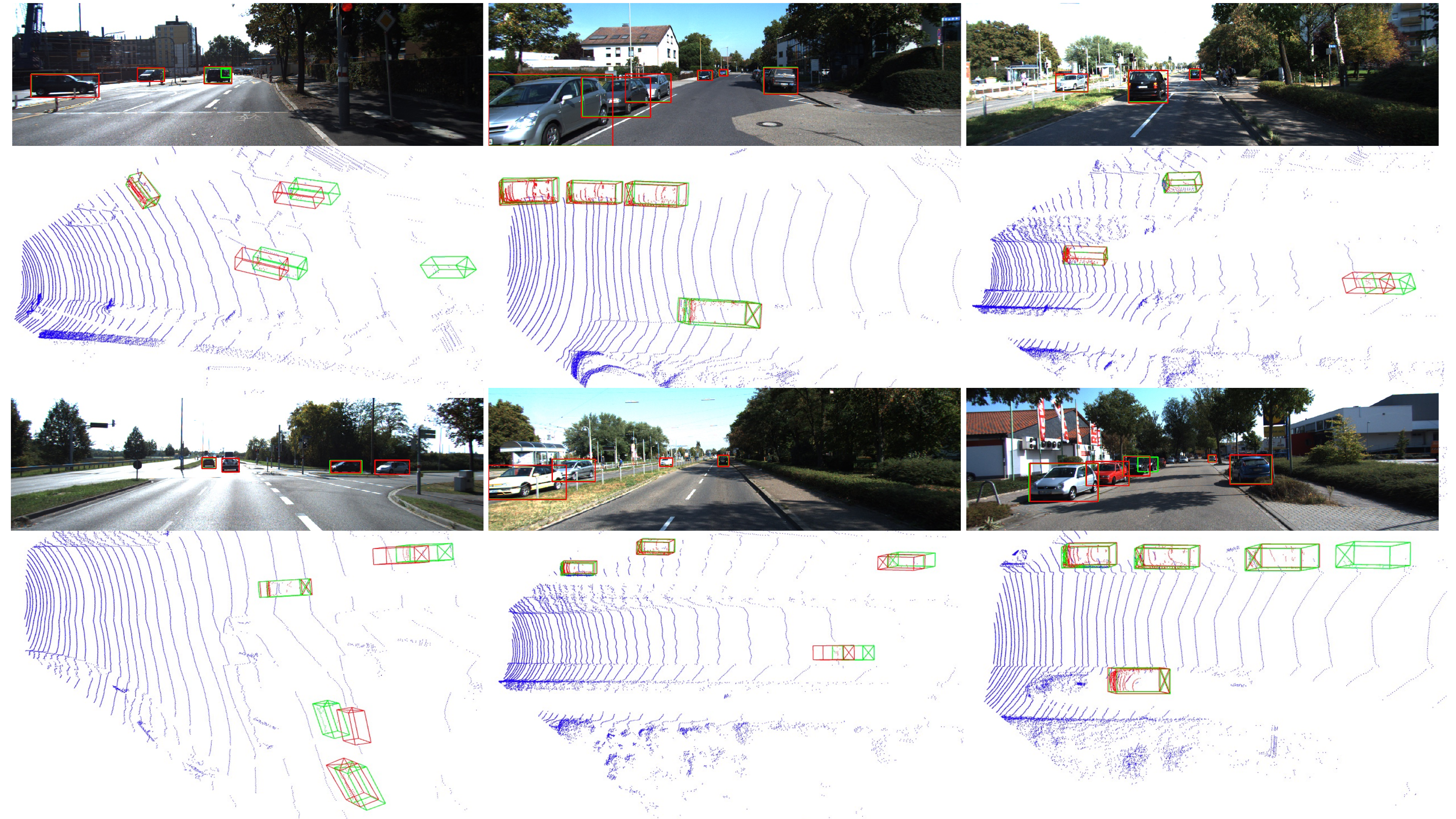}
\caption{Qualitative results on the KITTI val set. LiDAR point clouds are plotted for reference but not used in our method. We use \textcolor[RGB]{0,255,0}{green} and \textcolor[RGB]{255,0,0}{red} to denote predictions and ground truths, respectively.}
\label{Fig: visualize}
\end{figure*}

To visually evaluate the performance of our method based on GUPNet, we illustrate some examples in Figure~\ref{Fig: visualize}.
To clearly show the position of objects in the 3D world space, we also visualize the LiDAR signals and the ground-truth 3D boxes. 
We can observe that our outputs are remarkably accurate for the cases at a reasonable distance.
Unfortunately, it remains a challenge for occluded and truncated objects, a common dilemma for most monocular 3D detectors.

%-------------------------------------------------------------------------
\section{Limitations and Future Work}
Although our work tries to alleviate the effect of the depth ambiguity problem, the prediction of depth in monocular images is still an ill-posed problem. The occluded and truncated objects which drop some pixel information are even more challenging to be detected. Our OBMO module only allows the network to learn a reasonable depth range, making depth prediction more flexible. And it can not improve the confidence of an object. If an object with a low 3D object score, it is still difficult to know its depth range. We will consider the above situations in future work.
%-------------------------------------------------------------------------
\section{Conclusion}
In this paper, we point out that it is hard to predict depth accurately due to the enormous 3D space. According to this discovery, we design a simple but elegant plug-and-play module OBMO. We add pseudo labels under the X-Z ratio and Y-Z ratio, and design two kinds of label scores: IoU Label Score and Linear Label Score. Compared with existing monocular 3D object detection methods, OBMO achieves better performance on challenging KITTI and Waymo benchmarks.\looseness=-1

\begin{table*}[tbp]\centering
\caption{Experimental results of the Car category on the Waymo Open Dataset Validation Set.}
\begin{tabular}{@{}c|c|c|cccc@{}}
\toprule
\multirow{2}{*}{ Difficulty } & \multirow{2}{*}{  Threshold  } & \multirow{2}{*}{  Method } &  \multicolumn{4}{c}{3D mAP / 3D mAPH}\\
& & & \text{  Overall  } & \text{  $0 - 30$ m  } & \text{  $30 - 50$ m  } & \text{  $50 - \infty$ }\\
\hline\hline\noalign{\smallskip}
\multirow{12}{*}{LEVEL 1} & \multirow{6}{*}{IoU=0.7} & PatchNet & \text{0.39/0.37} & \text{1.67/1.63} & \text{0.13/0.12} & \text{0.03/0.03}\\
& & CADDN~\cite{CaDDN} & 5.03/4.99 & 14.54/14.43 & 1.47/1.45 & 0.10/0.10\\
& & PCT~\cite{PCT} & 0.89/0.88 & 3.18/3.15 & 0.27/0.27 & 0.07/0.07\\
& & MonoJSG~\cite{MonoJSG} & 0.97/0.95 & 4.65/4.59 & 0.55/0.53 & 0.10/0.09\\
 & & GUPNet~\cite{GUPNet} & 8.00/7.94 & 22.71/22.54 & 3.17/3.15 & 0.39/0.38\\
& & (+ OBMO) & \textbf{8.64/8.56} & \textbf{23.39/23.17} & \textbf{4.50/4.47} & \textbf{0.52/0.52}\\
\cline{2-7} 
& \multirow{6}{*}{IoU=0.5} & PatchNet & 2.92/2.74 & 10.03/9.75 & 1.09/0.96 & 0.23/0.18\\
& & CADDN~\cite{CaDDN}& 17.54/17.31 & 45.00/44.46 & 9.24/9.11 & 0.64/0.62\\
& & PCT~\cite{PCT} & 4.20/4.15 & 14.70/14.54 & 1.78/1.75 & 0.39/0.39\\
& & MonoJSG~\cite{MonoJSG} & 5.65/5.47 & 20.86/20.26 & 3.91/3.79 & 0.97/0.92\\
& & GUPNet~\cite{GUPNet} & 17.52/17.37 & 43.95/43.59 & 11.33/11.24 & \textbf{1.04/1.03}\\
& & (+ OBMO) & \textbf{20.71/20.53} & \textbf{48.08/47.64} & \textbf{16.97/16.85} & 0.68/0.68\\
\midrule
\multirow{12}{*}{LEVEL 2} & \multirow{6}{*}{IoU=0.7} & PatchNet & \text{0.38/0.36} & \text{1.67/1.63} & \text{0.13/0.11} & \text{0.03/0.03}\\
& & CADDN~\cite{CaDDN}& 4.49/4.45 & 14.50/14.38 & 1.42/1.41 & 0.09/0.09\\
& & PCT~\cite{PCT} & 0.66/0.66 & 3.18/3.15 & 0.27/0.26 & 0.07/0.07\\
& & MonoJSG~\cite{MonoJSG} & 0.91/0.89 & 4.64/4.65 & 0.55/0.53 & 0.09/0.09\\
& & GUPNet~\cite{GUPNet} & 7.57/7.51 & 22.64/22.47 & 3.10/3.08 & 0.36/0.36\\
& & (+ OBMO) & \textbf{8.28/8.20} & \textbf{23.33/23.12} & \textbf{4.42/4.40} & \textbf{0.51/0.51}\\
\cline{2-7} 
& \multirow{6}{*}{IoU=0.5} & PatchNet & \text{2.42/2.28} & \text{10.01/9.73} & \text{1.07/0.94} & \text{0.22/0.16}\\
& & CADDN~\cite{CaDDN}& 16.51/16.28 & 44.87/44.33 & 8.99/8.86 & 0.58/0.55\\
& & PCT~\cite{PCT} & 4.03/3.99 & 14.67/14.51 & 1.74/1.71 & 0.36/0.35\\
& & MonoJSG~\cite{MonoJSG} & 5.34/5.17 & 20.79/20.19 & 3.79/3.67 & 0.85/0.82\\
& & GUPNet~\cite{GUPNet} & 16.41/16.28 & 43.80/43.44 & 10.99/10.91 & \textbf{0.90/0.90}\\
& & (+ OBMO) & \textbf{19.41/19.23} & \textbf{47.91/47.47} & \textbf{16.46/16.35} & 0.59/0.59\\
\bottomrule
\end{tabular}
\label{Tab: waymo}
\end{table*}

\section*{Acknowledgments}
This work was supported in part by The National Nature Science Foundation of China (Grant Nos: 62036009, 62273302, 62273303, 62303406, 61936006), in part by Ningbo Key R\&D Program (No.2023Z231, 2023Z229), in part by the Key R\&D Program of Zhejiang Province, China (2023C01135), in part by Yongjiang Talent Introduction Programme (Grant No: 2022A-240-G, 2023A-194-G).
We sincerely thank Liang Peng, Minghao Chen, Menghao Guo for their suggestion in writing and Zhou Yang for the color scheme on the figures.\looseness=-1

% \newpage

\bibliographystyle{IEEEtran}
\bibliography{OBMO}
% \end{thebibliography}
 
\newpage

\vspace{11pt}
%  \bf{If you include a photo:}\vspace{-33pt}
\begin{IEEEbiography}[{\includegraphics[width=1in,height=1.25in,clip,keepaspectratio]{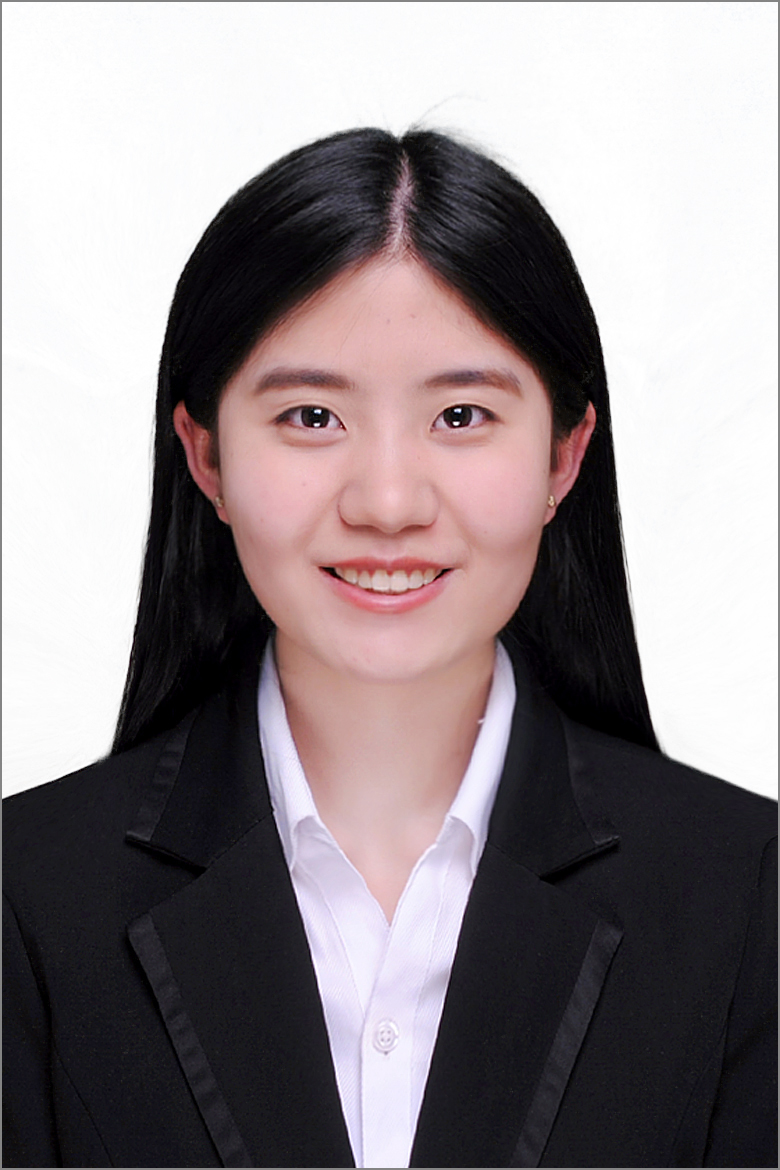}}]{Chenxi Huang} is a Ph.D. candidate supervised by Prof. Deng Cai in the State Key Lab of CAD\&CG, College of Computer Science at Zhejiang University, China. Her research interests include computer vision  and 3D scene understanding.
\end{IEEEbiography}

\vspace{11pt}
\begin{IEEEbiography}[{\includegraphics[width=1in,height=1.25in,clip,keepaspectratio]{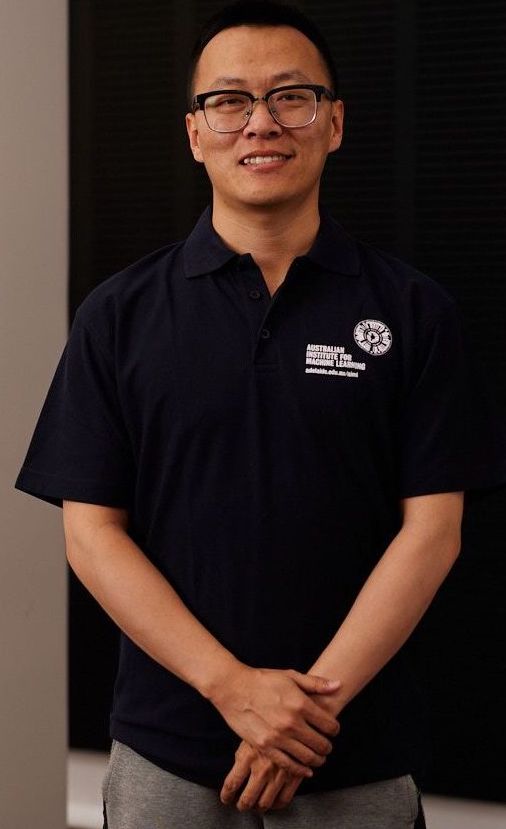}}]{Tong He} received the Ph.D. degree in computer science from the University of Adelaide, Australia, in 2020. He is currently a researcher at Shanghai AI Laboratory. His research interests include computer vision and machine learning.
\end{IEEEbiography}

\vspace{11pt}
\begin{IEEEbiography}[{\includegraphics[width=1in,height=1.25in,clip,keepaspectratio]{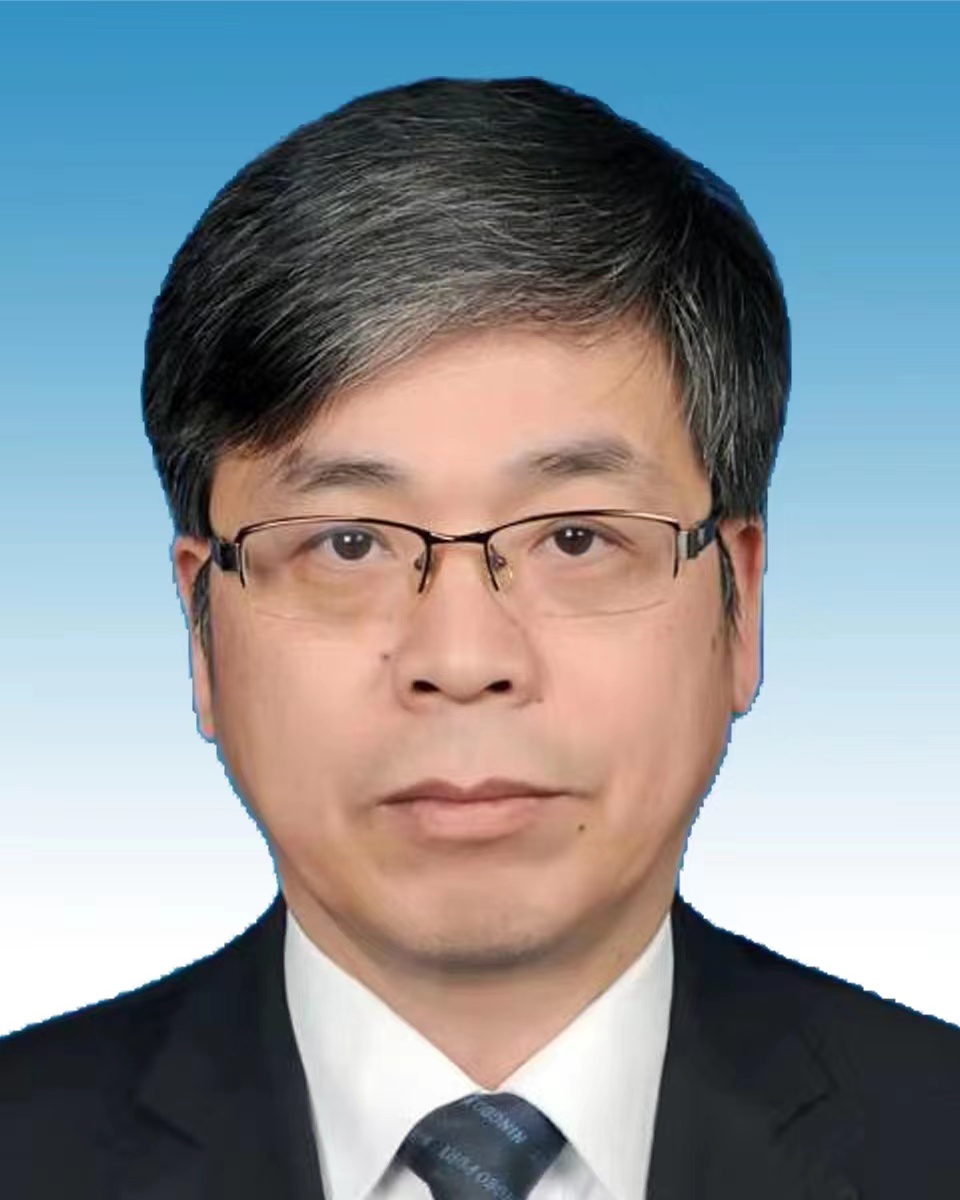}}]{Haidong Ren} is the deputy chief engineer of Ningbo Zhoushan Port Group Co.,Ltd, Ningbo, China. He is the director of Technology and Information Management Department, the director of the Science and Technology Center, a senior engineer, a post-doctoral supervisor, and a member of the Digital Reform Expert Group of the Zhejiang SASAC. He has long been engaged in the research and practice of port equipment and port informatization.
\end{IEEEbiography}

\vspace{11pt}
\begin{IEEEbiography}[{\includegraphics[width=1in,height=1.25in,clip,keepaspectratio]{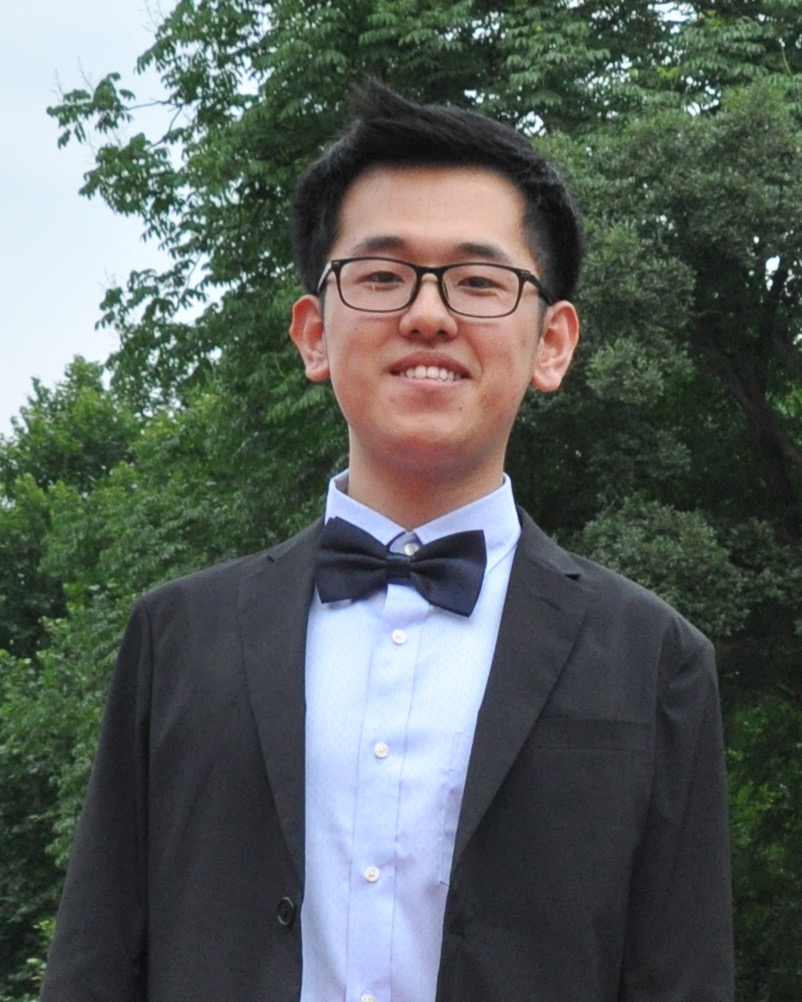}}]{Wenxiao Wang} is an assistant professor, School of Software Technology at Zhejiang University, China. He received the Ph.D. degree in computer science and technology from Zhejiang University in 2022. His research interests include deep learning and computer vision.
\end{IEEEbiography}

\vspace{11pt}
\begin{IEEEbiography}[{\includegraphics[width=1in,height=1.25in,clip,keepaspectratio]{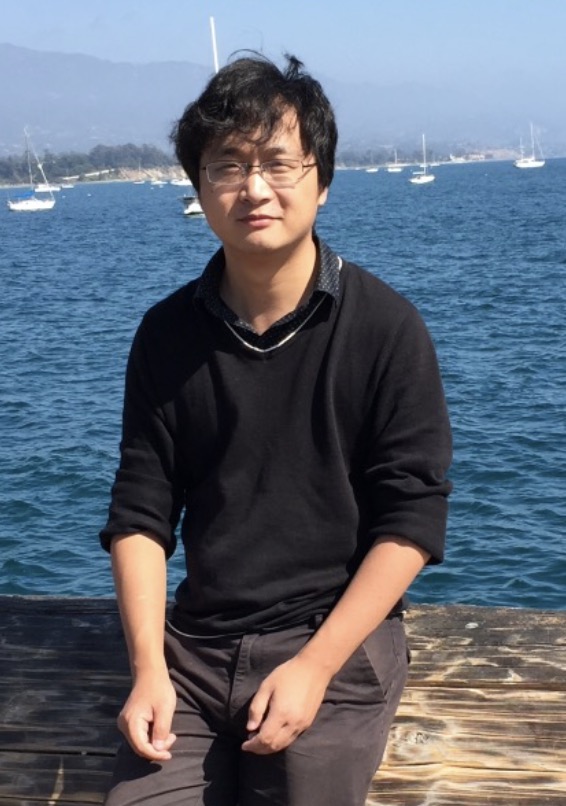}}]{BinBin Lin} is an assistant professor in the School of Software Technology at Zhejiang University, China. He received a Ph.D degree in computer science from Zhejiang University in 2012. His research interests include machine learning and decision making.
\end{IEEEbiography}

\vspace{11pt}
\begin{IEEEbiography}[{\includegraphics[width=1in,height=1.25in,clip,keepaspectratio]{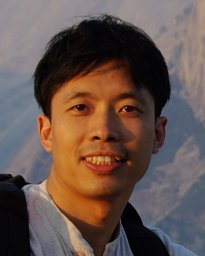}}]{Deng Cai} is a Professor in the State Key Lab of CAD\&CG, College of Computer Science at Zhejiang University, China. He received a Ph.D. degree in computer science from the University of Illinois at Urbana Champaign in 2009. His research interests include machine learning, data mining and information retrieval.
\end{IEEEbiography}

% Use the author name as the argument followed by the biography text.
 
\vfill

\end{document}